%% file: main.tex
\documentclass[10pt,journal, compsoc]{IEEEtran}
\usepackage{amsmath,amssymb,amsfonts}
\usepackage{textcomp}
\usepackage{url}
\usepackage{float}
\usepackage[table,xcdraw]{xcolor}
\usepackage{threeparttable}
\usepackage{times}
\usepackage{epsfig}
\usepackage{graphicx}
\usepackage{multirow}
\usepackage{algorithm}
\usepackage{algpseudocode}
\usepackage{amsmath}
\usepackage{amssymb}
\usepackage{pifont}
\usepackage{subfigure}
\usepackage{bbding}
\definecolor{mygray}{gray}{.9}

\usepackage{expl3}
\ExplSyntaxOn
\newcommand\latinabbrev[1]{
  \peek_meaning:NTF . {
    #1\@}%
  { \peek_catcode:NTF a {
      #1.\@ }%
    {#1.\@}}}
\ExplSyntaxOff

\ifCLASSOPTIONcompsoc
  \usepackage[nocompress]{cite}
\else
  \usepackage{cite}
\fi

\begin{document}
%

\title{UniSOT: A Unified Framework for Multi-Modality Single Object Tracking}
\author{Yinchao Ma,
        Yuyang Tang,
        Wenfei Yang,
        Tianzhu Zhang\footnotemark{*},
        Xu Zhou,
        and~Feng Wu
    \IEEEcompsocitemizethanks
    {\IEEEcompsocthanksitem *Corresponding Author. Yinchao Ma, YuYang Tang, Wenfei Yang, Tianzhu Zhang, Feng Wu are with the School of Information Science, University of Science and Technology of China, Hefei 230027, China
    (e-mail:
    imyc@mail.ustc.edu.cn;
    yuyangtang@mail.ustc.edu.cn,
    tzzhang@ustc.edu.cn;
    zhyd73@ustc.edu.cn;
    fengwu@ustc.edu.cn).
    Xu Zhou is with the Sangfor Research Institute.
    }
}

\markboth{IEEE Transactions on Pattern Analysis and Machine Intelligence,~Vol. XX, ~No. XX,~Aug 2024}
{YANG \MakeLowercase{\textit{et al.}}: Uncertainty Guided Collaborative Training for Weakly Supervised and Unsupervised Temporal Action Localization}

\IEEEtitleabstractindextext{%
\begin{abstract}
\input{abst}
\end{abstract}
}

\maketitle
\IEEEdisplaynontitleabstractindextext

\IEEEraisesectionheading{\section{Introduction}}
\label{sec:introduction}
\input{intro}

\section{Related Work}
\label{sec:related work}
\input{rw}

\section{Our Method}
\label{sec:method}
\input{our}

\section{Experiment}
\label{sec:experiment}

\input{expr}

\section{Conclusion}
\label{sec:conclusion}
\input{conclusion}

\section{Supplementary Materials}
\label{sec:sup}

\input{sup}

\ifCLASSOPTIONcaptionsoff
  \newpage
\fi

\bibliographystyle{IEEEtran}
\bibliography{main}

\end{document}

%% file: abst.tex
Single object tracking aims to localize target object with specific reference modalities (bounding box, natural language or both) in a sequence of specific video modalities (RGB, RGB+Depth, RGB+Thermal or RGB+Event.).
Different reference modalities enable various human-machine interactions, and different video modalities are demanded in complex scenarios to enhance tracking robustness.
Existing trackers are designed for single or several video modalities with single or several reference modalities, which leads to separate model designs and limits practical applications.
Practically, a unified tracker is needed to handle various requirements.
To the best of our knowledge, there is still no tracker that can perform tracking with these above reference modalities across these video modalities simultaneously.
Thus, in this paper, we present a unified tracker, UniSOT, for different combinations of three reference modalities and four video modalities with uniform parameters.
Extensive experimental results on 18 visual tracking, vision-language tracking and RGB+X tracking benchmarks demonstrate that UniSOT shows superior performance against modality-specific counterparts.
Notably, UniSOT outperforms previous counterparts by over 3.0\% AUC on TNL2K across all three reference modalities and outperforms Un-Track by over 2.0\% main metric across all three RGB+X video modalities.

%

%% file: intro.tex
Single object tracking focuses on localizing target object using specific reference modalities (bounding box, natural language or both) within sequences of particular video modalities (RGB, RGB+Depth, RGB+Thermal or RGB+Event).
Different reference modalities provide various human-machine interactions~\cite{UVLTrack}.
Different video modalities are demanded in complex scenarios to enhance robustness~\cite{protrack}.
Over the years, great progress has been achieved in single object tracking, which has spawned a wide range of applications in robotics, autonomous driving, human-machine interaction and so on~\cite{vottask3}.
Existing trackers can be divided into different categories by reference and video modalities.

\begin{figure*}[t]
    \centering
    \includegraphics[width=0.9\linewidth]{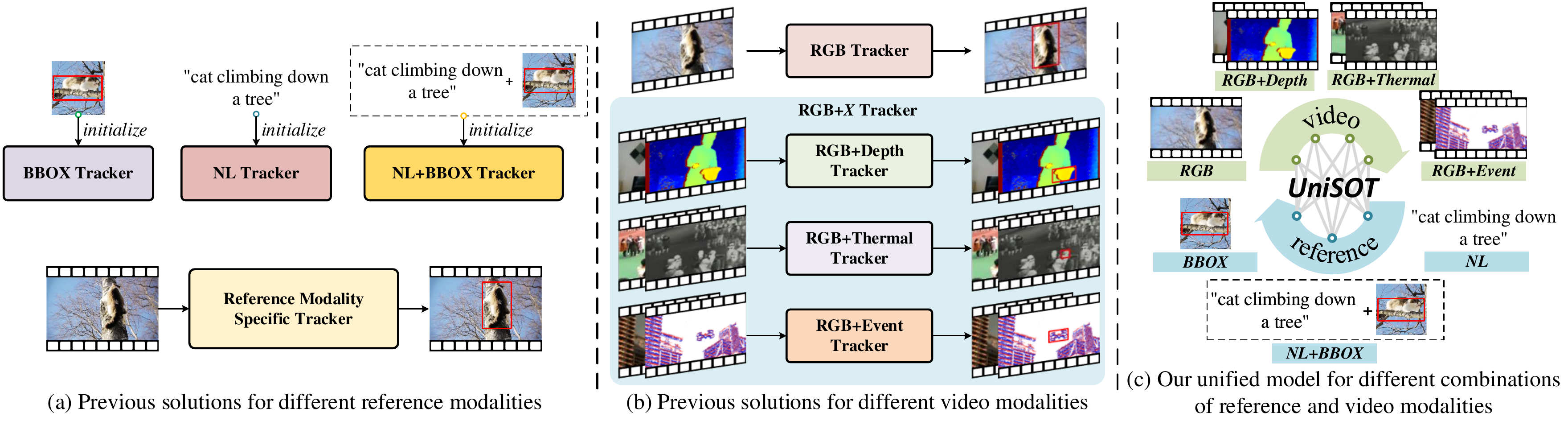}
    \vspace{-3mm}
    \caption{Comparison between previous solutions and UniSOT. 
    There are various reference and video modalities in single object tracking tasks for different application scenarios.
    However, previous trackers are commonly tailored for a specific reference or video modality.
    (a) BBOX, NL, NL+BBOX trackers utilize the bounding box(BBOX), natural language(NL), or both(NL+BBOX) as target object reference to track on RGB sequences.
    (b) RGB, RGB+Depth, RGB+Thermal, RGB+Event trackers are designed for tracking in sequences of corresponding video modality with BBOX reference.
    (c) Unlike the customized solutions for specific reference or video modalities, we seek to design a unified tracker (UniSOT) which can perform tracking with different combinations of three reference modalities and four video modalities using uniform parameters, enabling generalized capability.
    }
    \label{fig:introduction}
    \vspace{-4mm}
\end{figure*}

Based on \textit{reference modalities}, trackers can be divided into three categories, as shown in Figure~\ref{fig:introduction}(a).
\textbf{1) Bounding box} (\textbf{BBOX}). Most trackers~\cite{zhang2013robust,zhang2018robust,yan2021learning} utilize the target bounding box as the reference.
They commonly crop a template according to the given bounding box and localize target object in subsequent frames by interacting with the cropped template.
%
%
\textbf{2) Natural language} (\textbf{NL}). Different from the above tracking paradigm, tracking by natural language specification~\cite{nltrack} provides a novel manner of human-machine interaction.
This task first localizes target object in the first frame based on the language description, and then tracks target object based on the language description and the predicted bounding box~\cite{JointNLT}.
%
%
\textbf{3) Natural language and bounding box} (\textbf{NL+BBOX}).
For providing more accurate target reference, some trackers~\cite{VLT,SNLT} specify target object by both natural language and bounding box.
They embed language descriptions into visual features through dynamic filter~\cite{nltrack}, cross-correlation~\cite{SNLT} or channel-wise attention~\cite{VLT}, achieving more robust tracking.

Based on \textit{video modalities}, mainstream trackers can be divided into two categories, as shown in Figure~\ref{fig:introduction}(b).
\textbf{1)~RGB}. Traditional trackers perform tracking on RGB sequences~\cite{SiameseFC,DiMP}, which contain rich appearance cues to discriminate target object.
\textbf{2)~RGB+X}. \textit{X} means auxiliary video modality (depth, thermal or event). Recently, auxiliary modalities are drawing increasing attention in tracking tasks due to their complementary information with RGB images~\cite{ViPT}.
Most RGB+X trackers~\cite{rgbe1,rgbt1,depthtrack} integrate well-designed fusion modules to enhance RGB features with one specific auxiliary video modality ($e.g.$ RGB+Event), achieving more robust results in complex scenarios.

Previous reference or video modality-specific trackers lead to separate model designing, as shown in Figure~\ref{fig:introduction}.
In practice, users may desire various reference modalities for convenience~\cite{nltrack}, while the demand for video modalities varies across scenarios~\cite{protrack}.
%
%
It brings out the need for a unified tracker that can handle different combinations of reference and video modalities simultaneously.
Some trackers attempt to unify partial reference modalities~\cite{JointNLT,UVLTrack} or video modalities~\cite{UnTrack,protrack}. 
For example, JointNLT~\cite{JointNLT} can perform tracking with NL or NL+BBOX references on RGB sequences, and Un-Track can perform tracking with BBOX references on RGB+Depth, RGB+Thermal and RGB+Event sequences.
However, there is still no tracker that can perform tracking with three reference modalities across four video modalities.
%
The difficulties in designing such a unified tracker come from multiple aspects.
From the reference modality aspect, the semantic gap across reference modalities brings the difficulty of consistent feature learning and stable object localization~\cite{VLT}.
From the video modality aspect, it is difficult to learn video odality-aligned features across video modalities while retaining their specific cues~\cite{UnTrack}.

By studying previous methods, we summarize two main issues that need to be considered to build a unified single object tracker.
\textbf{1)~How to design a tracking model for various reference modalities.}
Recent trackers~\cite{ViPT,UVLTrack} follow a neat but effective architecture with a feature extractor and a box head.
The semantic gap across reference modalities poses a challenge in consistent feature learning for the feature extractor and stable object localization for the box head.
Thus, on the one hand, we need to \textit{design a generalized feature extractor for different references.}
Recent trackers introduce natural language reference through well-designed fusion modules, such as channel-wise attention~\cite{VLT} and Transformer blocks~\cite{JointNLT}.
However, they ignore the semantic gap between different modalities, resulting in a tendency for these trackers to rely on semantic information in language references, which limits their generalization on bounding box references~\cite{VLT}.
To this end, it is necessary to design a new reference-generalized feature extractor via vision-language alignment, achieving consistent feature learning for different reference modalities.
On the other hand, we need to \textit{design a robust box head across reference modalities.}
Existing trackers design various box heads to estimate target object state, such as anchor-free head~\cite{OSTrack}, corner-based head~\cite{JointNLT}, and point-based head~\cite{apmt}.
These heads commonly take the reference-enhanced features of the search region as input, and regress the target box through offline trained parameters.
However, enhanced features of search region vary with reference modalities,while the box head processes all search regions identically despite variations in reference modality, which may lead to unstable results across reference modalities.
Thus, we argue that it is better to design a reference-adaptive head, which can localize target object dynamically according to reference modalities for stable tracking.
%
2)~\textbf{How to design a training strategy for various video modalities.}
%
%
Recently, some RGB+X trackers~\cite{ViPT, UnTrack} fine-tune extra low-rank modules on frozen foundation models, avoiding overfitting on small RGB+X datasets while retaining foundational tracking capabilities.
\textcolor{black}{However, they commonly project auxiliary modalities into different low-rank spaces with an identical rank and fuse with RGB features.
Such a design poses a challenge for video modality-aligned feature learning, and ignores the potential difference in the amount of information across video modalities that may desire various ranks.
Thus, we seek to design a video modality-adaptive adaptation mechanism for video modality joint tuning, achieving both video modality-aligned and video modality-specific feature learning.
}

Motivated by the above discussions, we propose a unified tracking framework, termed UniSOT, with two designing modules for reference and video modality unification, respectively.
\textbf{For reference modality designing}, we propose a reference-generalized feature extractor and a reference-adaptive box head, which are trained on large-scale RGB tracking datasets with different reference modalities.
The \textit{reference-generalized feature extractor} is a new architecture constructed based on Transformer~\cite{2017Attention}, in which we extract features of vision and language separately in shallow encoder layers and fuse them in deep encoder layers.
Such a design avoids the confusion of low-level feature modeling between different reference modalities and allows high-level semantics interaction.
Besides, we design a multi-modal contrastive loss to align visual and language features into a unified semantic space, so as to realize consistent feature learning for different reference modalities.
The \textit{reference-adaptive box head} dynamically mines scenario information from video contexts by reference features of different modalities and localizes target object in a contrastive way for stable tracking.
Specifically, we propose a novel distribution-based cross-attention mechanism, which can adaptively mine features of target, distractor and background from historical frames by different reference modalities.
Then, target object can be localized directly through feature comparison.
\textbf{For video modality designing}, we freeze parameters trained on RGB tracking datasets with different reference modalities and then introduce incremental parameters to train with auxiliary video modalities.
Specifically, we design a \textit{rank-adaptive modality adaptation} (RAMA) mechanism inspired by AdaLoRA~\cite{adalora}, which evaluates the importance of incremental parameters from video modality-shared and video modality-specific perspectives, and dynamically adjusts the parameter ranks for different video modalities.
Thus, RAMA can learn not only video modality-aligned features through shared parameters but also video modality-specific features via adaptive ranks, enabling UniSOT to perform robust tracking with different auxiliary video modalities in a uniform parameter set.
Extensive experimental results on 18 tracking benchmarks with various reference and video modalities demonstrate that UniSOT shows superior performance against reference or video modality-specific counterparts.
%

To summarize, the main contributions of this work are:
(1) We propose a novel unified tracking framework for different combinations of reference and video modalities. 
To the best of our knowledge, UniSOT is the first tracker to support tracking with three reference modalities (NL, BBOX, NL+BBOX) in sequences of four video modalities (RGB, RGB+Depth, RGB+Thermal, RGB+Event) with uniform parameters, enabling more generalized application scenarios.
(2) To address the challenges in unified tracking with different reference modalities, we design a reference-generalized feature extractor with multi-modal contrastive loss based on Transformer for consistent feature learning, and a reference-adaptive box head to stabilize target localization via contrastive learning of dynamic scenarios.
(3) To address the challenges in unified tracking with different video modalities, we propose a rank-adaptive modality adaptation mechanism inspired by AdaLoRA for both video modality-aligned and video modality-specific feature learning, which enables various video modalities in UniSOT and provides a novel training paradigm for incremental video modalities.

%
%

%% file: rw.tex
%
In this section, we will introduce single object tracking methods based on different reference and video modalities.
\subsection{Trackers based on Different Reference Modalities}
Different reference modalities can provide information of target object from different aspects and enable various human-machine interactions. Based on the reference modalities, existing trackers can be divided into three categories, including BBOX, NL and NL+BBOX trackers.

\textbf{BBOX}. Most existing trackers localize target object in a video sequence according to the bounding box in the first frame~\cite{OTB2015}.
These trackers commonly crop a target template from the first frame and estimate target object state in subsequent frames by interacting with the cropped template.
Two-stream trackers~\cite{SiamRPNplusplus,apmt,ATOM} extract features from both template and search region and then localize target object through building feature interactions between them by cross-correlation layer~\cite{SiamBAN,SiamCAR,SiameseFC}, correlation filter~\cite{xu2021adaptive,ECO,DiMP} or attention blocks~\cite{yan2021learning,TransT,AiATrack}.
Recently, one-stream trackers~\cite{cui2022mixformer,OSTrack} achieve joint feature extraction and interaction using Transformer~\cite{wu2021cvt,2017Attention}, which simplify the tracking pipeline and achieve superior performance.

\begin{figure*}[t]
    \centering
    \includegraphics[width=0.85\linewidth]{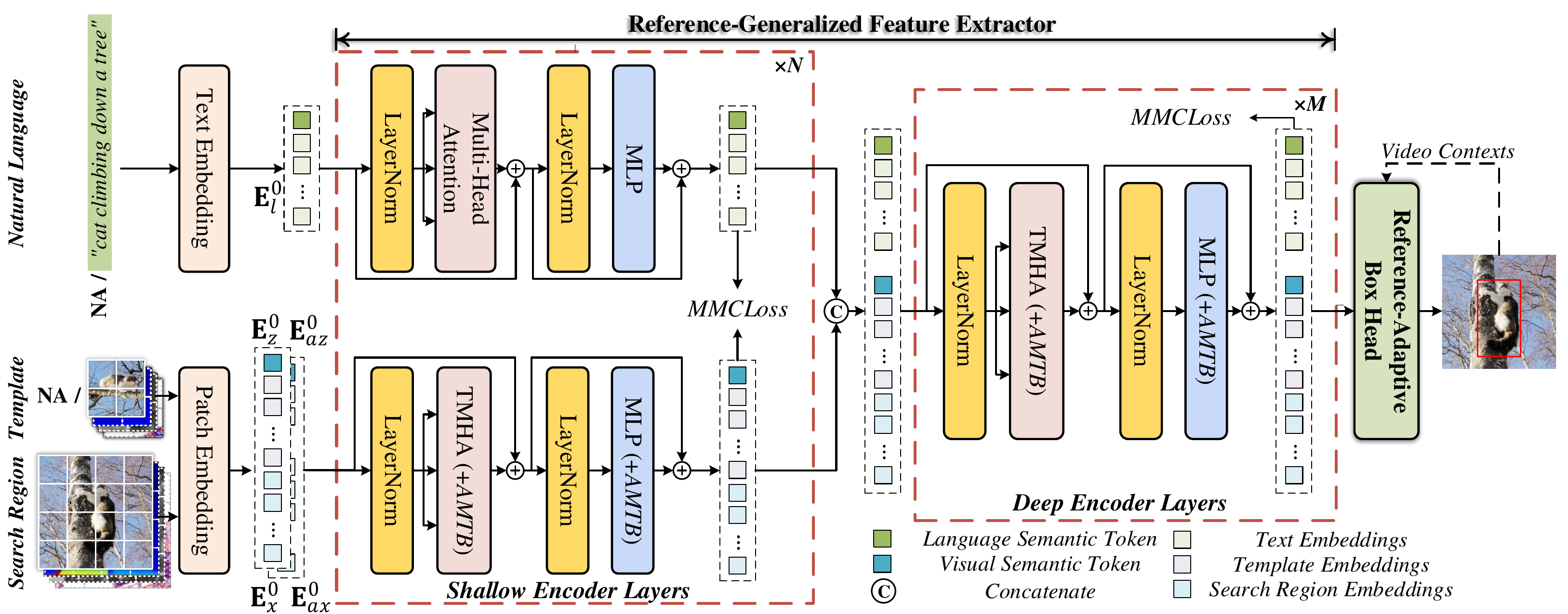}
    \vspace{-3mm}
    \caption{
    A unified tracking framework for different target references.
    {NA} means ``not available", which is filled with zeros. Natural language is not available for visual tracking task, and the template is not available for grounding task.
    Different from previous trackers designed for specific reference or video modalities, our UniSOT can simultaneously handle different combinations of reference and video modalities in a unified framework.
    }
    \label{fig:architecture}
    \vspace{-4mm}
\end{figure*}

\textbf{NL}. Different from the above tracking paradigm, some trackers~\cite{nltrack,TSN,TNL2K} specify target object purely based on natural language, which provides a novel human-computer interaction manner.
Li \textit{et al.}~\cite{nltrack} first define this task and provide a baseline by combining a grounding model and a tracking model.
Then, some trackers~\cite{TSN,TNL2K,GTI} follow this paradigm to design different models to solve the grounding task and tracking task separately.
Recently, some trackers~\cite{JointNLT,querynlt} perform tracking and grounding using a unified model, which simplifies the overall tracking framework and enables end-to-end optimization.

\textbf{NL+BBOX}. 
To provide more accurate target references, some trackers~\cite{nltrack,VLT} specify target object through both the initial bounding box and natural language.
Li \textit{et al.}~\cite{nltrack} first introduce natural language into tracking achieving more robust results than visual trackers, which demonstrates the potential of vision-language tracking.
SNLT~\cite{SNLT} embeds natural language into Siamese-based trackers as a convolutional kernel and localizes target object through cross-correlation.
VLT~\cite{VLT} treats the natural language feature as a selector to weigh different visual feature channels, enhancing target-related channels for robust tracking.
Also, JointNLT~\cite{JointNLT} introduces natural language by interacting language and visual features in Transformer blocks.
%
%
However, due to the semantic gap between vision and language, these trackers trained with natural language show limited performance with bounding box (BBOX) reference~\cite{VLT}.
To this end, we design a multi-modal contrastive loss to align features of different reference modalities into a unified semantic space for consistent feature learning.
Meanwhile, a reference-adaptive box head is proposed to alleviate the instability of target localization across reference modalities.
Thanks to effective designs, UniSOT achieves promising performance across all reference modalities with high FPS.
\subsection{Trackers based on Different Video Modalities}
\noindent
Different video modalities can provide complementary information for target localization. Based on the video modalities, mainstream trackers can be divided into two categories, including RGB and RGB+X trackers.

\textbf{RGB}. Most existing trackers~\cite{SiameseFC,apmt,SiamCAR} are designed for RGB video modality, which contains rich appearance cues to discriminate target object.
These trackers extract features for both the target template and search region, and then match features between them through correlation filter~\cite{ASRCF,CCOT,KCF,ECO}, convolution filter~\cite{song2017crest,ATOM,DiMP}, cross-correlation~\cite{SiameseFC,SiameseRPN,SiamRPNplusplus} or attention mechanism~\cite{TransT,yan2021learning,fu2021stmtrack} to localize target object.
Recent trackers~\cite{cui2022mixformer,OSTrack,SimTrack} allow interaction between target template and search region in the feature extraction phase for target-aware feature learning.
Although great progress has been achieved, these RGB trackers still have inherent shortcomings in challenging scenarios, which have few appearance cues such as occlusion, fast motion, low illumination and so on~\cite{ViPT}.

\textbf{RGB+X}.
Auxiliary video modalities (depth, thermal and event) are drawing increasing attention due to their complementary information with RGB images.
Traditionally, RGB-D trackers~\cite{depthtrack,rgbd1k}, RGB-T trackers~\cite{rgbt1,rgbt2}, RGB-E trackers~\cite{rgbe1,visevent} design video modality-specific modules to integrate auxiliary video modalities, achieving more robust performance under complex scenarios.
Recently, some works~\cite{protrack,ViPT} seek to adapt foundation RGB trackers for RGB-X tracking tasks via video modality-specific fine-tuning, which can make full use of the pretrained knowledge of the foundation model.
Furthermore, Un-Track~\cite{UnTrack} proposes a video modality-joint fine-tuning method via low-rank factorization and reconstruction techniques.
However, it projects different video modalities into different low-rank spaces with an identical rank.
Such a design poses a challenge for video modality-aligned feature learning, and ignores the potential difference in the amount of information across video modalities, which may desire various ranks.
To address the above limitations, we design a rank-adaptive modality adaptation (RAMA) mechanism for video modality-joint tuning, which can dynamically adjust the rank of shared parameters for different auxiliary video modalities, enabling both video modality-shared and video modality-specific feature learning.

%% file: our.tex
In this section, we first introduce the overall architecture of UniSOT, which presents a simple but effective tracking pipeline for different combinations of reference and video modalities.
The following two subsections introduce reference and video modality designing respectively.

\subsection{Tracking Architecture}
As shown in Figure~\ref{fig:architecture}, UniSOT can perform tracking with different combinations of reference modalities (NL, BBOX, NL+BBOX) and video modalities (RGB, RGB+Depth, RGB+Thermal, RGB+Event).
The template is cropped based on the initial bounding box.
The search region is cropped based on the last-frame bounding box.
Given the language description $l$, we tokenize the sentence and embed each word with the text embedding layer to obtain text embeddings $\mathbf{E}_l^0\in\mathbb{R}^{N_l\times{C}}$.
$N_l$ is the maximum text length, $C$ is the feature dimension.
Given the RGB template $z\in\mathbb{R}^{3\times{H_z}\times{W_z}}$ with its auxiliary video modality $z_a\in\mathbb{R}^{3\times{H_z}\times{W_z}}$, the RGB search region (full image for grounding) $x\in\mathbb{R}^{3\times{H_x}\times{W_x}}$ with its auxiliary video modality $x_a\in\mathbb{R}^{3\times{H_x}\times{W_x}}$, we split and reshape them into a sequence of flattened 2D patches and then project them into a latent space.
Learnable position embeddings are added to the corresponding patch embeddings, obtaining template embeddings $\mathbf{E}_z^0\in\mathbb{R}^{N_z\times C}$ with its auxiliary modality embeddings $\mathbf{E}_{az}^0\in\mathbb{R}^{N_z\times C}$, search region embeddings $\mathbf{E}_x^0\in \mathbb{R}^{N_x\times C}$ with its auxiliary video modality embeddings $\mathbf{E}_{ax}^0\in \mathbb{R}^{N_x\times C}$.
Here, the position embeddings for the template and search region are independent to implicitly reflect their roles as different image types (template vs. search region), while those for different video modalities associated with the same image type are shared to align their position information.
$N_z$ and $N_x$ are the patch numbers of the template and search region respectively.
Here, the unavailable (\textbf{NA}) features for different reference and video modalities will be filled with zeros.
We also prepend a language semantic token $\mathbf{T}_l^0\in \mathbb{R}^{1\times C}$ and a visual semantic token $\mathbf{T}_v^0\in \mathbb{R}^{1\times C}$ to text embeddings and image embeddings correspondingly, which are designed to capture the global semantics of different reference modalities.
After that, text and image embeddings are fed into the reference-generalized feature extractor, which is built on Transformer architecture.
Specifically, we extract language and visual features separately in shallow encoder layers and fuse them in deep encoder layers.
A multi-modal contrastive loss (MMCLoss) is proposed to align vision and language features into a unified semantic space.
Moreover, features of RGB modality and auxiliary video modality are interacted in auxiliary modality tuning blocks (AMTB).
Finally, we feed the interacted text and image embeddings into the reference-adaptive box head, which can make full use of different reference modalities to mine ever-changing scenario features from video contexts and localize target object in a contrastive way.
Further, search region embeddings with high-confidence target bounding boxes are saved as video contexts $\mathbf{E}_c^{N+M}$ to help with subsequent target localization.

\subsection{Reference Modality Designing}

\subsubsection{Reference-Generalized Feature Extractor}
For unified tracking with different reference modalities, three critical issues must be carefully addressed in feature extraction: multi-modal feature learning and fusion, reference modality-compatible training, and multi-modal feature alignment.
Thus, we design a new architecture for reference modality generalized feature extraction, termed reference-generalized feature extractor.
%
As shown in Figure~\ref{fig:architecture}, the reference-generalized feature extractor is constructed based on Transformer~\cite{2017Attention}, which consists of $N$ shallow encoder layers and $M$ deep encoder layers.
Here, ``shallow" and ``deep" refer to the sequential position of the layers in the reference-generalized feature extractor.
We extract visual and language features separately in shallow encoder layers and fuse them in deep encoder layers, which can avoid the confusion of low-level feature modeling for different reference modalities, and allow high-level semantics interaction for target localization.

\begin{figure}[t]
    \centering
    \includegraphics[width=1.0\linewidth]{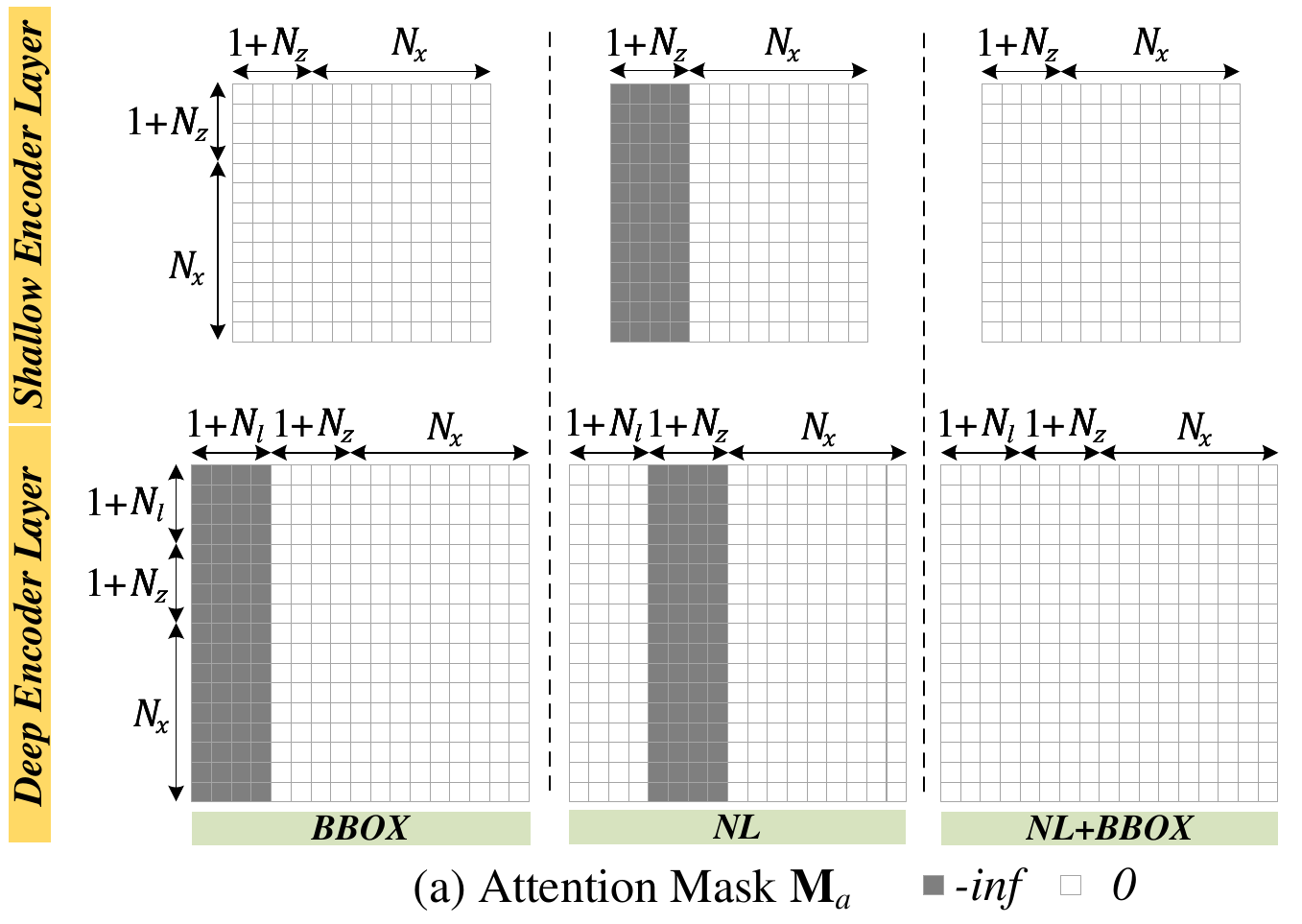}
    \vspace{-5mm}
    \caption{
    The attention mask of task-oriented multi-head attention for different target references. It enables different reference modalities to be trained jointly.
    }
    \label{fig:attention_mask}
    \vspace{-2mm}
\end{figure}

For modality-compatible training of different reference inputs, we fill the unavailable reference embeddings with zeros and propose a task-oriented multi-head attention mechanism (TMHA) to avoid task-irrelevant feature interactions, inspired by masking mechanism in Transformer~\cite{2017Attention}.
For simplicity, we present the single-head formulas of TMHA below.
Given the input of $i^{th}$ encoder layer $\mathbf{E}^{i-1}$, key ${\mathbf{K}}^i$, query ${\mathbf{Q}}^i$ and value $\mathbf{V}^i$ arise from $\mathbf{E}^{i-1}$ through layer normalization and linear projections.
Then, we filter task-irrelevant feature interactions in attention mechanisms through masking. The output of $i^{th}$ encoder layer can be formulated as,
%
\begin{gather}
    \widehat{\mathbf{E}}^i =
    {\rm Softmax}(\frac{\mathbf{Q}^i(\mathbf{K}^i)^\top}{\sqrt{C}}+\textbf{M}_a)\mathbf{V}^i+\mathbf{E}^{i-1},\\
    \mathbf{E}^{i} = {\rm MLP}({\rm LN}(\widehat{\mathbf{E}}^i)) + \widehat{\mathbf{E}}^i,
\end{gather}
where ${\rm MLP(\cdot)}$ is multi-layer perception, ${\rm LN(\cdot)}$ is layer normalization, $\widehat{\mathbf{E}}^i$ is a intermediate variable, $\mathbf{E}^{i}$ is the output of $i^{th}$ encoder layer. $\textbf{M}_a$ is the attention mask, which is related to the input reference type.
Figure~\ref{fig:attention_mask} shows the details of the attention mask for different reference modalities, where we fill the positions of unavailable features with $-\inf$ and other positions with $0$.

\begin{figure}[t]
    \centering
    \includegraphics[width=0.9\linewidth]{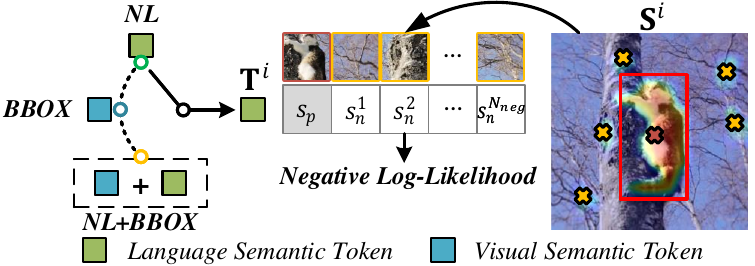}
    \vspace{-3mm}
    \caption{
    The diagram of the multi-modal contrastive loss. We align different reference modalities with target object patch feature by contrasting hard background patches.
    }
    \label{fig:loss}
    \vspace{-3mm}
\end{figure}

%
Furthermore, during joint training with different reference modalities, we find that semantic tokens derived from different reference modalities exhibit significant discrepancies on the corresponding maps of the search region and demonstrate limited discriminability to the target object. 
This indicates that, within a unified framework, features from different reference modalities suffer from a substantial semantic gap and poor discriminative power, ultimately hindering their effectiveness in robust tracking.
%
%
Thus, inspired by cross-modal alignment algorithms~\cite{radford2021learning,li2021align}, we specifically design a simple yet effective multi-modal contrastive loss (MMCLoss) to enhance the similarity between visual/language semantic tokens and target features in the search region while suppressing similarity with background distractors ($e.g.,$ top $N_{neg}$ scores out of target object box).
By this design, we can align vision and language features into a unified semantic space and improve the discriminability of reference features.
%
As shown in Figure~\ref{fig:loss}, we first acquire the semantic token $\mathbf{T}^i$ of $i^{th}$ encoder layer according to the reference modalities (NL, BBOX or NL+BBOX), which ensures compatibility with different reference modalities during joint training.
\begin{gather}
    \mathbf{T}^i = 
    \begin{cases}
    \mathbf{T}^i_l & NL, \\
    \mathbf{T}^i_v & BBOX, \\
    (\mathbf{T}_l^i + \mathbf{T}_v^i)/2 & NL+BBOX.
    \end{cases}
    \label{equ:T}
\end{gather}
Given the semantic token $\mathbf{T}^i$ of $i^{th}$ encoder layer, we compute the similarity $\mathbf{S}^i=[s^{i,1};s^{i,2},...,s^{i,N_x}]$ between $\mathbf{T}^i$ and search region embeddings $\mathbf{E}_x^i=[f^{i,1};f^{i,2},...,f^{i,N_x}]$.
Formally,
\begin{small}
\begin{gather}
    s^{i,j} = {\rm sim}(\mathbf{T}^i, f^{i,j}) / \tau,
    {\rm sim}(\mathbf{T}^i, f^{i,j}) = \frac{\mathbf{T}^i(f^{i,j})^\top}{||\mathbf{T}^i||_2||f^{i,j}||_2},
\end{gather}
\end{small}

\noindent
where $\tau$ is a temperature parameter, $||\cdot||_2$ means $l_2$ norm.
According to $\mathbf{S}^i$, we select the central score of target object $s_p^i$ as a positive sample score and top $N_{neg}$ scores out of target object box $[s_n^{i,k}]_{k=1}^{N_{neg}}$ as negative sample scores.
Finally, the multi-modal contrastive loss can be formulated as follows,
\begin{gather}
    \mathcal{L}_{mmc}^i = -{\rm log}(\frac{e^{s_p^i}}{e^{s_p^i}+\sum_{k=1}^{N_{neg}}e^{s_n^{i,k}}}).
\end{gather}

\noindent
\textbf{Discussion}. 
%
%
Unlike prior image-wise or video-wise contrastive learning losses~\cite{radford2021learning,li2021align,yang2021taco}, which focus on text-image or text-video holistic alignment, MMCLoss is a target-wise contrastive loss specifically tailored for unified tracking with diverse reference modalities.
It focuses on the alignment between different reference modalities and target features, thereby aligning visual and language features into a unified semantic space and enhancing the discriminability of reference features for robust tracking.
%

\begin{figure*}[t]
    \centering
    \includegraphics[width=0.98\linewidth]{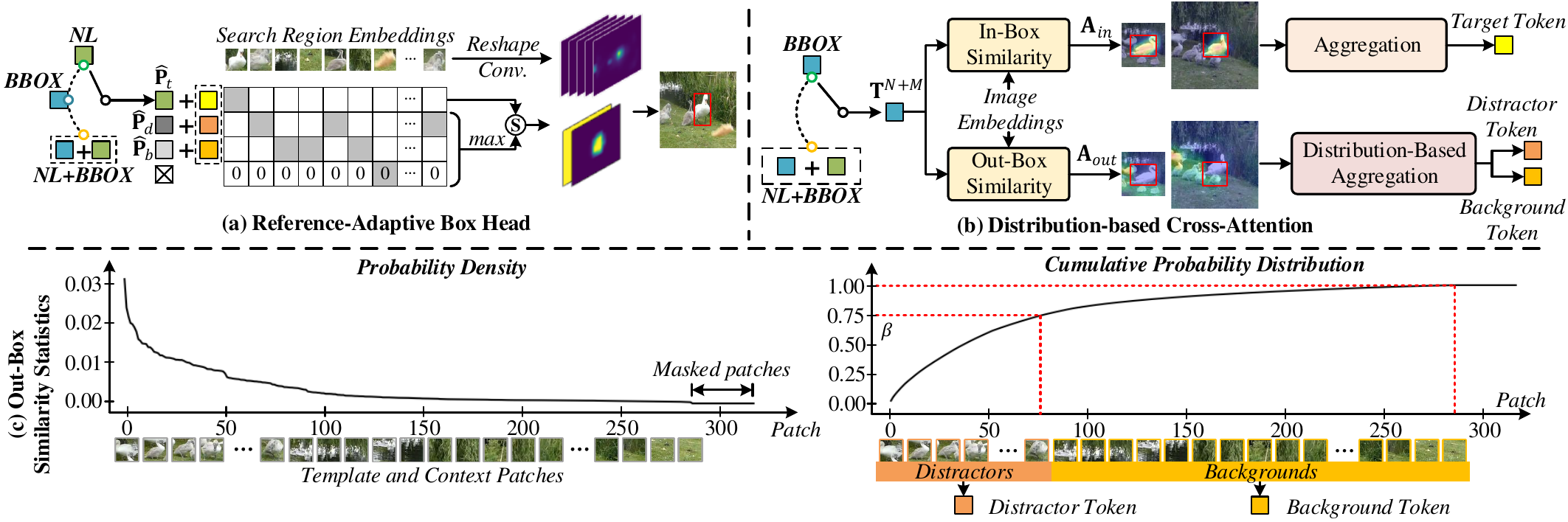}
    \vspace{-3mm}
    \caption{(a) shows the schematic of the reference-adaptive box head, which can make full use of reference information to discriminate target object. (b) shows the structure of the distribution-based cross-attention. (c) illustrates the descending order of similarity between the semantic token and template/context background patches, which can be regarded as the probability of background features being mistakenly classified as the target object. The cumulative probability distribution clarifies how we apply the threshold $\beta$ to separate distractor features from background features.}
    \label{fig:head}
    \vspace{-4mm}
\end{figure*}

\subsubsection{Reference-Adaptive Box Head}

Inspired by OSTrack~\cite{OSTrack}, we reshape the reference enhanced embeddings of search region $\textbf{E}_x^{N+M}$ into a 2D feature map and feed it into a three-branch convolutional network to regress a center score map $\hat{\mathbf{C}}\in (0,1)^{\frac{H_x}{p}\times\frac{W_x}{p}}$, an offset map $\hat{\mathbf{O}}\in [0,1)^{2\times\frac{H_x}{p}\times\frac{W_x}{p}}$ and a normalized box size map $\hat{\mathbf{S}}\in (0,1)^{2\times\frac{H_x}{p}\times\frac{W_x}{p}}$, where $p$ is the size of image patches.
%
%
However, enhanced features of search region vary with reference modalities, while the box head processes all search regions identically despite variations in reference modality, which may lead to unstable results across reference modalities.
%
Thus, we further design a reference-adaptive box head (RABH), which can adaptively mine ever-changing scenario features from video context by reference features of different modalities to assist target localization for robust tracking.

As shown in Figure~\ref{fig:head}(a), we treat the semantic token in the last encoder layer as the target prototype $\widehat{\mathbf{P}}_t=\mathbf{T}^{N+M}$ (as defined in Equation~\ref{equ:T}) and introduce learnable distractor prototype $\widehat{\mathbf{P}}_d$ and background prototype $\widehat{\mathbf{P}}_b$ to localize target object in a contrastive way.
Here, $\widehat{\mathbf{P}}_d\in\mathbb{R}^{1\times{C}}$ and $\widehat{\mathbf{P}}_b\in\mathbb{R}^{1\times{C}}$ are shared across reference modalities, which provide a general capability to discriminate backgrounds when video contexts are unavailable, $e.g.$, localizing the target object in the first frame for natural language reference (NL).
During tracking, video contexts can provide rich scenario cues to discriminate target object.
Thus, as shown in Figure~\ref{fig:head}(b), we propose a novel distribution-based attention mechanism to mine features of the target, distractor and background from historic frames.
Historic search region embeddings with high-confidence target bounding boxes are saved as context embeddings $\mathbf{E}_c^{N+M}$ (as stated in Section~3.1).
Given the template and context embeddings $\mathbf{E}_t=[\mathbf{E}_z^{N+M};\mathbf{E}_c^{N+M}]\in\mathbb{R}^{(N_z+N_x)\times{C}}$ and the target masks $\textbf{M}_t=[\textbf{M}_z;\textbf{M}_c]\in\mathbb{R}^{{1}\times(N_z+N_x)}$, we compute in-box similarity $\mathbf{A}_{in}$ and out-box similarity $\mathbf{A}_{out}$ between the target semantic token $\mathbf{T}^{N+M}$ and $\mathbf{E}_t$ to obtain the probability that the patch belongs to the target. Formally,
\begin{gather}
    \mathbf{A}_{in} = {\rm Softmax}(\frac{\mathbf{T}^{N+M}\mathbf{E}_t^\top}{\sqrt{C}}+\textbf{M}_t),\\
    \mathbf{A}_{out} = {\rm Softmax}(\frac{\mathbf{T}^{N+M}\mathbf{E}_t^\top}{\sqrt{C}}+\widetilde{\textbf{M}}_t),
\end{gather}
where $\textbf{M}_t$ is obtained by assigning the position in the target box to 0 and the position out of the target box to $-inf$. 
$\widetilde{\textbf{M}}_t$ is the opposite.
Then, the target token $\mathbf{T}_t$ is obtained by in-box similarity aggregation $\mathbf{T}_t = \mathbf{A}_{in}\mathbf{E}_t$.
Considering that the distractor with a similar appearance to target object is a key factor affecting the tracking robustness~\cite{KeepTrack}, we divide features out of the target box into distractor and background from the perspective of distribution.
As shown in Figure~\ref{fig:head}(c), we rank the out-box probabilities $\mathbf{A}_{out}$ in descending order and sum them cumulatively to obtain the cumulative probability distribution.
The probability density can be regarded as the probability of background features being mistakenly classified as the target object.
We can find that some background patches have a similar appearance to the target object, which tends to confuse the tracker.
Thus, we separate these distractor features from background features using a threshold $\beta$ on the cumulative probability distribution to obtain distractor patches in the search region.
Specifically, we assign the patch whose target cumulative probability distribution score is lower than $\beta$ to 0 for $\textbf{M}_d$ and other positions to $-inf$.
$\widetilde{\textbf{M}}_d$ is the opposite.
Then, the distractor token and background token can be formulated as,
%
\begin{gather}
    \mathbf{T}_d = {\rm Softmax}(\frac{\mathbf{T}^{N+M}\mathbf{E}_t^\top}{\sqrt{C}}+\widetilde{\textbf{M}}_t+\textbf{M}_d)\mathbf{E}_t,\\
    \mathbf{T}_b = {\rm Softmax}(\frac{\mathbf{T}^{N+M}\mathbf{E}_t^\top}{\sqrt{C}}+\widetilde{\textbf{M}}_t+\widetilde{\textbf{M}}_d)\mathbf{E}_t.
\end{gather}
After obtaining $\mathbf{T}_t,\mathbf{T}_d,\mathbf{T}_b$, we add them to scenario prototypes $\widehat{\mathbf{P}}_t,\widehat{\mathbf{P}}_d,\widehat{\mathbf{P}}_b$ to supplement the dynamic scenario information.
Given search region embeddings $\mathbf{E}_x^{N+M}=[f^1;f^2,...,f^{N_x}]$, we compute the corresponding target similarity $\hat{\mathbf{L}}=[{\alpha}^1_t,{\alpha}^2_t,...,{\alpha}^{N_x}_t]$ as follows,
%
\begin{gather}
    {\mathbf{P}}_t = \widehat{\mathbf{P}}_t + \mathbf{T}_t,
    {\mathbf{P}}_d = \widehat{\mathbf{P}}_d + \mathbf{T}_d,
    {\mathbf{P}}_b = \widehat{\mathbf{P}}_b + \mathbf{T}_b,\\
    \hat{\alpha}^i_t = {\rm sim}({f^i, {\mathbf{P}}_t})/\tau, \\
    \hat{\alpha}^i_b = {\rm max}\big({\rm sim}(f^i,{\mathbf{P}}_d)/\tau,{\rm sim}(f^i,{\mathbf{P}}_b)/\tau,0\big),\\
    {\alpha}^i_t = {e^{\hat{\alpha}^i_t}}/{(e^{\hat{\alpha}^i_t}+e^{\hat{\alpha}^i_b})}.
\end{gather}
Here, we append a zero to the background score computation, which avoids unseen objects having relatively high target score $\alpha^i_t$ after the softmax operation.
Finally, given the position $(x_c,y_c)={\rm argmax}_{(x,y)}\hat{\mathbf{C}}(x,y)\hat{\mathbf{L}}(x,y)$, the bounding box of target $\hat{b}=(\hat{x},\hat{y},\hat{w},\hat{h})$ can be formulated as,
\begin{gather}
    (\hat{x},\hat{y}) = \Big(\big(x_c+\hat{\mathbf{O}}(0,x_c,y_c)\big)\cdot p,
                             \big(y_c+\hat{\mathbf{O}}(1,x_c,y_c)\big)\cdot p\Big), \\
    (\hat{w},\hat{h}) = \big(\hat{\mathbf{S}}(0,x_c,y_c)\cdot H_x, \hat{\mathbf{S}}(1,x_c,y_c)\cdot W_x\big).
\end{gather}

\subsubsection{Training Objective}
We train UniSOT on RGB datasets with different reference modalities, which have rich data sources, enabling powerful foundational tracking capabilities.
We call this the first training stage.
Specifically, we treat patches in the target box as positive samples and others as negative samples to generate the groundtruth ${\mathbf{L}}$ for target score map $\hat{\mathbf{L}}$.
Then, the binary cross-entropy loss is adopted for the target score map constraint.
Formally,
\begin{gather}
    \mathcal{L}_{tgt} = \mathcal{L}_{bce}(\hat{\mathbf{L}}, {\mathbf{L}})
\end{gather}
The training objectives of center score map $\mathcal{L}_{cls}$ and bounding box $\mathcal{L}_{box}=\lambda_1\mathcal{L}_1+\lambda_{giou}\mathcal{L}_{giou}$ are consistent with OSTrack~\cite{OSTrack}.
In summary, the overall objective function can be formulated as,
\begin{gather}
    \mathcal{L} = \mathcal{L}_{tgt}+\mathcal{L}_{cls}+\mathcal{L}_{box}+\lambda_{mmc}\sum_{i=1}^{N+M}\mathcal{L}_{mmc}^{i}.
\end{gather}

\input{method_extra}

%% file: method_extra.tex
\subsection{Video Modality Designing}
In the first training stage, UniSOT enables the unification of reference modalities on RGB sequences.
In this subsection, we design a rank-adaptive modality adaptation mechanism for video modality joint tuning, enabling UniSOT to freely integrate auxiliary video modalities for robust tracking.
\subsubsection{Rank-Adaptive Modality Adaptation}
Previous methods~\cite{UnTrack,ViPT} project auxiliary video modalities into different low-rank spaces with an identical rank and fuse with RGB features.
Such a design poses a challenge for video modality-aligned feature learning and ignores the potential difference in the amount of information across video modalities, which may desire various ranks.
To address the above limitation, we propose a novel solution for video modality joint tuning inspired by AdaLoRA~\cite{adalora}, enabling both video modality-aligned and video modality-specific feature learning.

\begin{figure}
    \centering
    \includegraphics[width=0.8\linewidth]{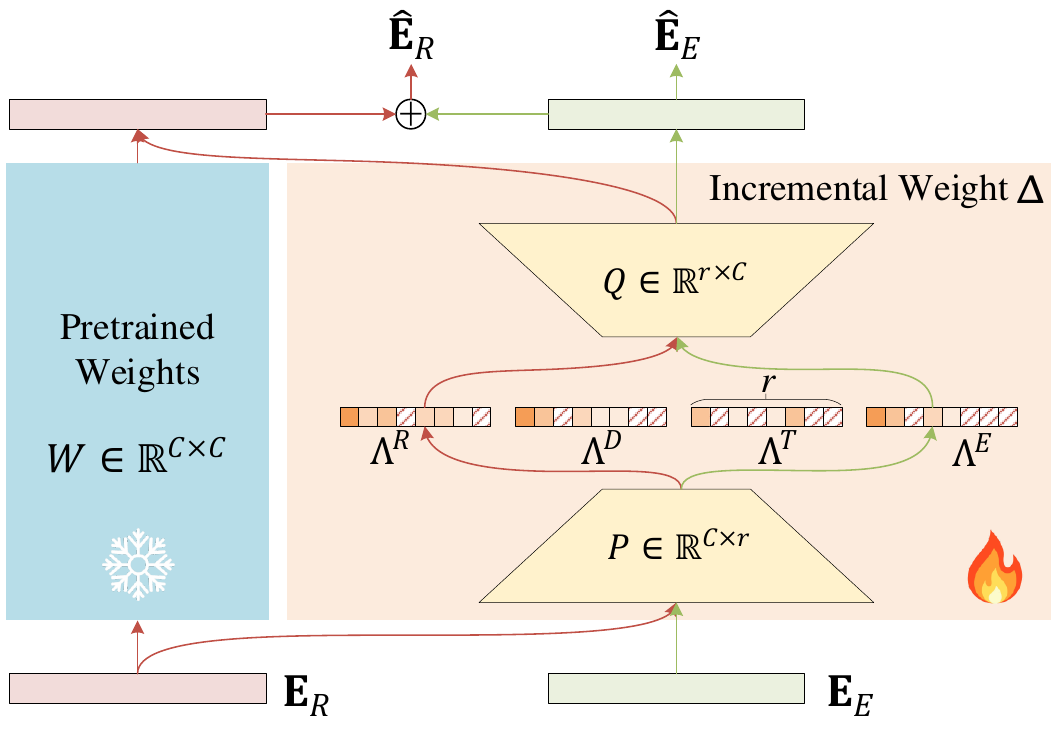}
    \caption{The diagram of auxiliary modality tuning block (AMTB). Different video modalities share a uniform parameter set $P$ and $Q$ for video modality-aligned feature learning. Meanwhile, we assign various ranks across video modalities via singular values $\Lambda$ to learn modality-specific cues.}
    \label{fig:rama}
\end{figure}

\begin{figure}
    \centering
    \includegraphics[width=\linewidth]{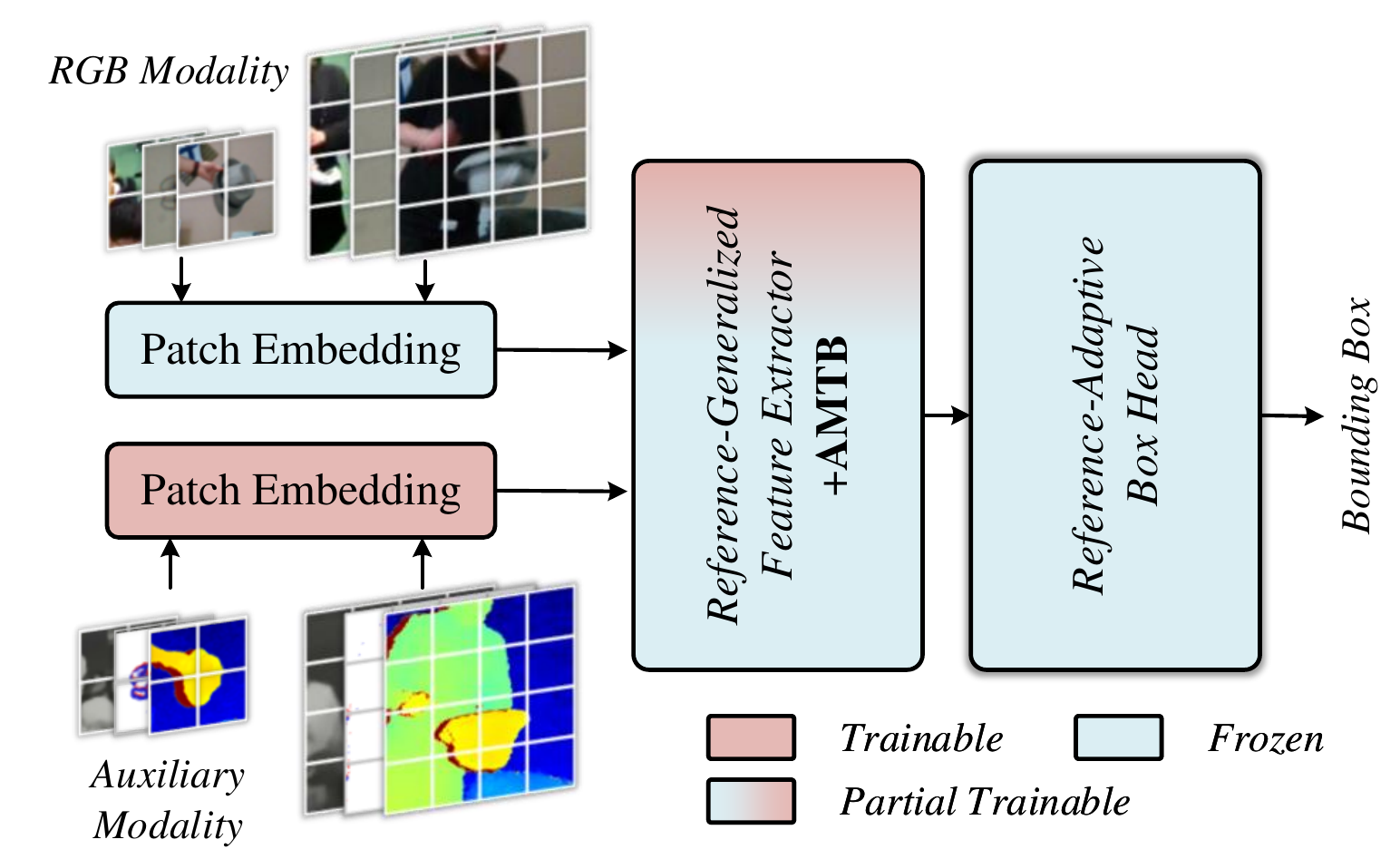}
    \vspace{-5mm}
    \caption{The diagram of UniSOT. UniSOT integrates auxiliary modality tuning blocks (AMTB) for modality-joint tuning, which allows UniSOT to unify RGB+Depth, RGB+Thermal, and RGB+Event tracking with a uniform parameter set.}
    \vspace{-3mm}
    \label{fig:uvltrackx}
\end{figure}

\noindent
\textbf{Formulation}. To empower the capability of auxiliary video modalities, we indirectly train some dense layers in UniSOT by parameterizing their adaptation changes as incremental weights, while freezing the pretrained weights.
Specifically, we design an auxiliary modality tuning block (AMTB) for video modality-joint tuning, as shown in Figure~\ref{fig:rama}, which can fuse auxiliary video modalities ($e.g.$ event feature $\mathbf{E}_E$) to RGB feature $\mathbf{E}_R$ by incremental weights $\Delta$.
AMTB can be injected into the feature extractor to integrate auxiliary video modalities, as shown in Figure~\ref{fig:uvltrackx}.
Inspired by AdaLoRA~\cite{adalora}, we parameterize the incremental weights $\Delta$ in the form of singular value decomposition $\Delta=P\Lambda Q$, where $P\in \mathbb{R}^{C\times r}$ and $Q\in \mathbb{R}^{r\times C}$ are orthogonal matrices, $\Lambda\in \mathbb{R}^{r\times r}$ contains singular values and $r$ denotes the inherent rank of $\Delta$.
In practice, since $\Lambda$ is diagonal, $\Lambda$ can be expressed as a vector in $\mathbb{R}^{r}$.
This weight format enables us to explicitly manipulate the rank of $\Delta$ by the number of non-zero elements in $\Lambda$.
Further, we assign different $\Lambda$ for different video modalities, termed $\Lambda^{R}$, $\Lambda^{D}$, $\Lambda^{T}$, $\Lambda^{E}$.
Here, $R$, $D$, $T$ and $E$ correspond to RGB, depth, thermal and event modalities, similarly hereinafter.
Different video modalities share a uniform parameter set $P$ and $Q$, which facilitates video modality-aligned feature learning.
Meanwhile, various $\Lambda$ allow networks to learn features with different ranks for different video modalities to obtain video modality-specific cues.
The output of the auxiliary modality tuning block can be formulated as,
\begin{gather}
    \hat{\mathbf{E}}_{a} = ReLU(\mathbf{E}_{a}P\Lambda^{a}{Q}),\\
    \hat{\mathbf{E}}_{R} = \mathbf{E}_{R}W+\mathbf{E}_{R}P\Lambda^{R}{Q}+\hat{\mathbf{E}}_{a},
\end{gather}
where $a\in \{D,T,E\}$ is the auxiliary video modality.

\begin{figure}
    \centering
    \includegraphics[width=\linewidth]{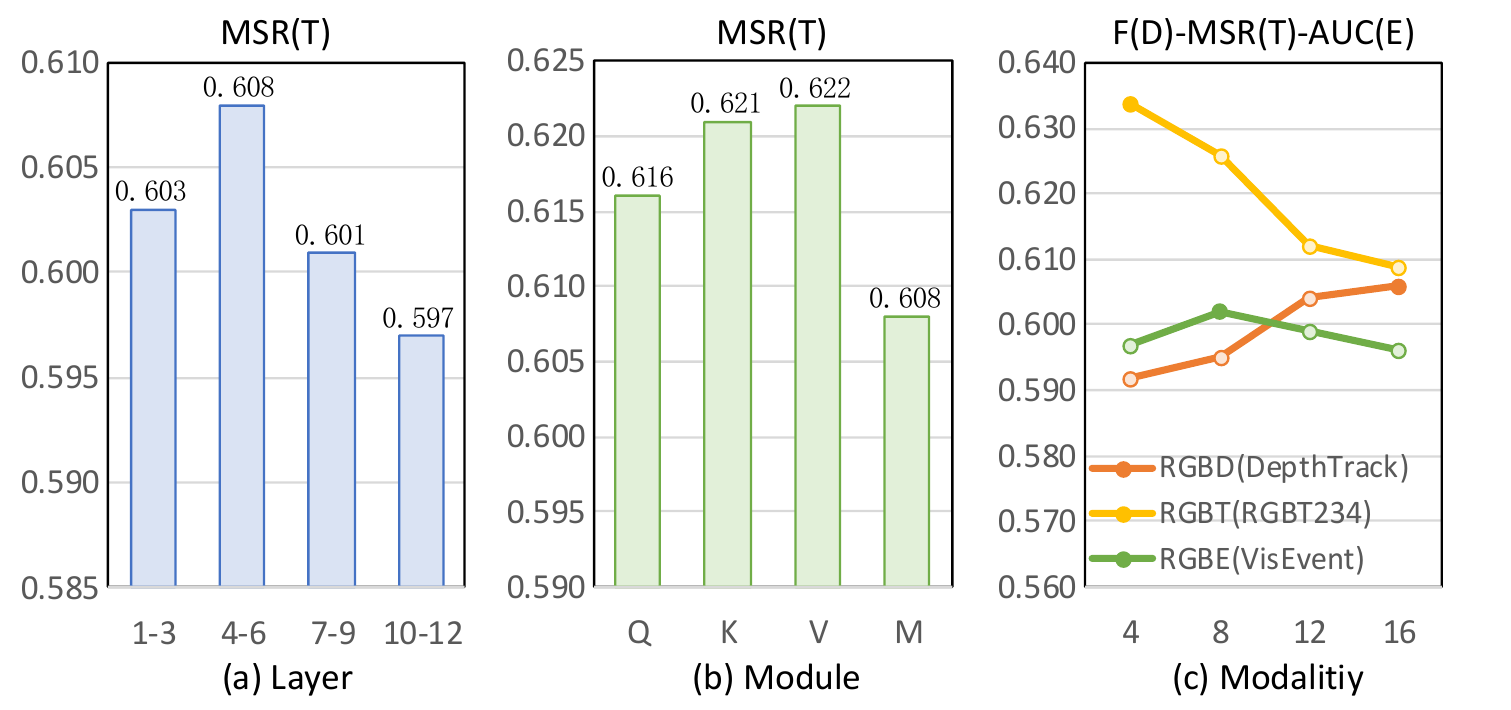}
    \caption{Tuning performance comparison across tuned layers, modules and modalities. We inject AMTB into UniSOT and fix elements of $\Lambda$s to~1. (a) and (b) show the tuning performance on RGBT234 dataset, which reflects the contribution gaps among incremental weights. (c) shows the performance for different video modalities and ranks, which indicates auxiliary video modalities desire various ranks to learn effective features. We adopt a common training configuration~\cite{ViPT} for these experiments with consistent epochs and sample numbers.}
    \label{fig:auxexp}
\end{figure}

\noindent
\textbf{Parameter Importance}.
As shown in Figure~\ref{fig:auxexp}, the performance of the model after fine-tuning varies with tuned layers, modules and modalities, which means the contribution gaps among incremental weights.
Also, as shown in Figure~\ref{fig:auxexp}(c), auxiliary video modalities desire various ranks to learn effective features, which poses a challenge for video modality-joint tuning.
To this end, we seek to evaluate the contribution of parameters and allocate different ranks across layers, modules and modalities.
So, we first define an importance function $s(\cdot)$ for incremental parameters:
\begin{equation}
    I(w_{ij})=|w_{ij}\nabla_{w_{ij}}\mathcal{L}|, \label{sensitivity}
\end{equation}
\begin{equation}
    \overline{I}^{(t)}=\beta_1\overline{I}^{(t-1)}+(1-\beta_1){I}^{(t)}, \label{sensitivity-1}
\end{equation}
\begin{equation}
    \overline{U}^{(t)}=\beta_2\overline{U}^{(t-1)}+(1-\beta_2)|\overline{I}^{(t)}-{I}^{(t)}|, \label{sensitivity-2}
\end{equation}
\begin{equation}
    s(w_{ij})=\overline{I}^{(t)}\cdot\overline{U}^{(t)}. \label{sensitivity-3}
\end{equation}
Equation~(\ref{sensitivity}) essentially measures the change of loss when $w_{ij}$ is zeroed out~\cite{sanh2020movement}. Equations~(\ref{sensitivity-1}--\ref{sensitivity-3}) further stabilizes $I(w_{ij})$ through cross-batch sensitivity smoothing and uncertainty quantification~\cite{zhang2022platon}, $t$ means iteration step.

\noindent
\textbf{Modality-shared Rank Allocation}.
For the sake of description, we further denote incremental weights $\Delta_k$ as parameter tuples $G_{k,i}=\{P_{k,*i},\lambda^{R}_{k,i},\lambda^{D}_{k,i},\lambda^{T}_{k,i},\lambda^{E}_{k,i},Q_{k,i*}\}$ containing the $i^{th}$ singular values and vectors in $k^{th}$ incremental weights.
Here, $P_{k,*i}\in\mathbb{R}^{C\times 1},Q_{k,i*}\in\mathbb{R}^{1\times C}$.
We first design a modality-shared importance $S_{k,i}$ for each parameter tuple, which is based on the importance of singular values and vectors. Formally,
\begin{gather}
\begin{split}
S_{k,i}=&\sum_{a\in \{R,D,T,E\}}s(\lambda_{k,i}^a)\\
    &+\frac{1}{d_1}\sum_{j=1}^{d_1}s(P_{k,ji})+\frac{1}{d_2}\sum_{j=1}^{d_2}s(Q_{k,ij}).
\end{split}
\end{gather}
Then, we reserve top-$n$ important singular values and zero out others according to $S_{k,i}$, adaptively constructing a compact modality-shared parameter space:
\begin{gather}
    \hat\lambda_{k,i}^a = 
	\left\{ \begin{array}{lc}
		\lambda_{k,i}^a & S_{k,i} \text{ is in the top-}n \text{ of } \mathcal{S}^1,\\
		0 & \text{ otherwise,}
	\end{array}
	\right.
\end{gather}
where $a\in\{R,D,T,E\}$ similarly hereinafter, $n$ denotes the modality-shared rank budget and $\mathcal{S}^1$ contains the importance scores of all parameter tuples $G_{k,i}$.

\noindent
\textbf{Modality-specific Rank Allocation}.
As shown in Figure~\ref{fig:auxexp}(c). Different video modalities desire various ranks to learn modality-specific cues and prevent overfitting.
We further split $\Delta_k$ as modality-specific parameter tuples $G^{a}_{k,i}=\{P_{k,*i},\lambda^{a}_{k,i},Q_{k,i*}\}$ and evaluate the modality-specific importance $S_{k,i}^{a}$ for each parameter tuple to adjust the rank of weight for different video modalities.
It can be formulated as,
\begin{gather}
    S_{k,i}^{a} = s(\hat\lambda_{k,i}^a)+\frac{1}{d_1}\sum_{j=1}^{d_1}s(P_{k,ji})+\frac{1}{d_2}\sum_{j=1}^{d_2}s(Q_{k,ij}).
\end{gather}
Similarly, we reserve top-$m$ important singular values and zero out others:
\begin{gather}
    \hat\lambda_{k,i}^a = 
	\left\{ \begin{array}{lc}
		\hat\lambda_{k,i}^a & S_{k,i}^{a} \text{ is in the top-}m \text{ of } \mathcal{S}^2,\\
		0 & \text{ otherwise,}
	\end{array}
	\right.
\end{gather}
where $m$ denotes the modality-specific rank budget and $\mathcal{S}^2$ contains the importance scores of all modality-specific parameter tuples $G^{a}_{k,i}$.
\subsubsection{Fine-tuning}
We further fine-tune UniSOT on RGB+X datasets, enabling various video modalities.
We call this the second training stage.
As shown in Figure~\ref{fig:uvltrackx}, auxiliary video modalities are fed to UniSOT together with RGB images (template and search region).
We freeze the parameters trained in the first training stage.
Specifically, we add a trainable patch embedding layer for feature extraction of auxiliary video modality.
Then, auxiliary modality tuning blocks are injected into the feature extractor for video modality-joint tuning.
To enforce the orthogonality of $P$ and $Q$, we add an orthogonal loss to the tuning objective $\mathcal{L}_{total}$ to enforce approximately orthogonal throughout training. $\mathcal{L}$ is consistent with the first stage. Formally,
\begin{gather}
    \mathcal{L}_{orth} = ||P^\top P-I||^2_F+||Q^\top Q-I||^2_F,\\
    \mathcal{L}_{total} = \mathcal{L} + \lambda_{orth}\mathcal{L}_{orth}.
\end{gather}

\noindent
\textbf{Discussion}.
Modality-shared rank allocation constructs a compact modality-shared parameter space conducive to modality-aligned feature learning.
Thanks to the feature alignment between RGB and auxiliary video modalities, these auxiliary video modalities can be integrated into vision-language tracking without any modification.
It enables UniSOT to simultaneously handle different combinations of reference and video modalities in a unified framework with a uniform parameter set.
Modality-specific rank allocation adaptively assigns different ranks across video modalities, allowing video modality-specific feature learning while avoiding overfitting.
Unlike AdaLoRA, RAMA introduces a cross-modal rank allocation strategy to address the learning problem of video modality-shared and video modality-specific features during video modality-joint tuning.
Meanwhile, RAMA is the first work to explore rank-adaptive allocation strategies in multi-modal joint fine-tuning, it empowers RGB-based models with multiple auxiliary video modalities through a single fine-tuning process, significantly streamlining training procedures. 
The resulting model enable multi-modal video tracking via a uniform parameter set, providing a reliable technical foundation for endowing RGB foundation models with multi-modal capabilities.

%% file: expr.tex
Our tracker is implemented using Python 3.8.13 and Pytorch 1.10.1. The UniSOT are trained on a server with eight 24GB NVIDIA RTX 3090 GPUs and AMD EPYC 7713 64-Core Processor @ 2GHz with 503 GB RAM.
The inference speed is tested on a single NVIDIA RTX2080 Ti GPU and Intel(R) Xeon(R) CPU E5-2695 v4 @ 2.10GHz with 251 GB RAM.
\subsection{Implementation Details}
\subsubsection{Network Details.}
We crop the template and search region by $2^2$ and $4^2$ times the target bounding box area and resize them to 128$\times$128 and 256$\times$256 respectively.
The test image for first frame grounding is scaled such that its long edge is 256.
Image patch size $p$=16.
%
For language, the max length of the sentence $N_l$ is set to 40.
To demonstrate the scalability, we present two variants, termed UniSOT-B and UniSOT-L.
The number of encoder layers is set to $N$=6,$M$=6 for UniSOT-B and $N=12$,$M=12$ for UniSOT-L.

\noindent
\subsubsection{The First Training Stage.}
\begin{figure}
    \centering
    \includegraphics[width=0.9\linewidth]{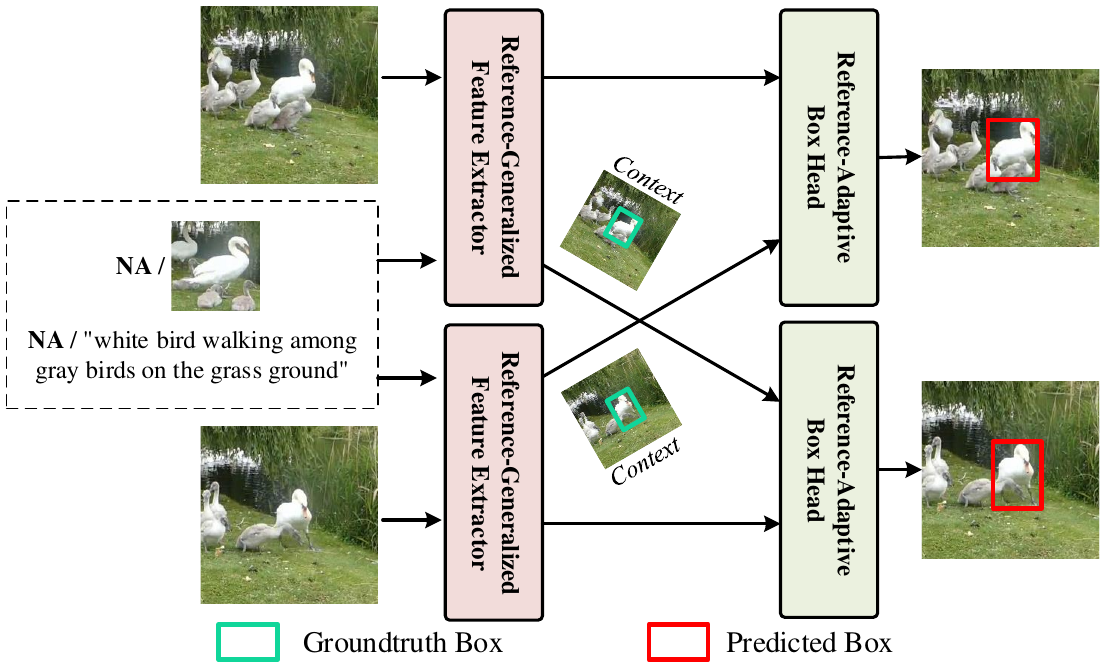}
    \vspace{-2mm}
    \caption{The diagram of the UniSOT training process. We sample two search regions from a video, whose features serve as context features for each other in the box head.}
    \vspace{-5mm}
    \label{fig:train}
\end{figure}

\textbf{Data Unit.} As shown in Figure~\ref{fig:train}, we sample one template, two search regions and the corresponding language description as a training unit in the first training stage.
Here, the template is not available for the natural language reference (NL).
Natural language is not available for the bounding box reference (BBOX).
The unavailable reference embeddings are filled with zeros in the reference-generalized feature extractor.
The training units for bounding box reference (BBOX) are sampled from the training splits of LaSOT~\cite{LaSOT}, GOT-10k~\cite{GOT10K}, COCO2017~\cite{COCO}, TrackingNet~\cite{trackingnet}, TNL2K~\cite{TNL2K}, OTB99~\cite{nltrack}.
The training units for natural language reference (NL) are sampled from the training splits of LaSOT, TNL2K, OTB99 and RefCOCOg-google~\cite{RefCOCOg}.
The training units for both bounding box and natural language reference (NL+BBOX) are sampled from the training splits of LaSOT, TNL2K, OTB and RefCOCOg-google.
The maximum sampling interval of search regions and target template is set to 200.
%
Common data augmentation is used for model training, such as translation, horizontal flip, and color jittering~\cite{yan2021learning}.
Due to the flexibility of our framework, different reference modalities can be trained jointly, which provides a neat training pipeline.

\noindent
\textbf{Initialization.}
The language branch in shallow encoder layers of UniSOT-B is initialized with uncased version parameters of BERT-Base~\cite{BERT}.
Other parameters in the reference-generalized feature extractor are initialized with ViT-Base parameters pretrained by MAE~\cite{he2022masked}.
Correspondingly, UniSOT-L parameters are initialized with BERT-Large and ViT-Large parameters.
The reference-adaptive box head is initialized with Xavier init~\cite{Xavier}.

\begin{table*}[!t]
\begin{center}
\caption{Comparison with state-of-the-art real-time visual trackers on TNL2K, AVisT, LaSOT, LaSOT$_{ext}$ and TrackingNet. The best two results are shown in bold and underline. 
}
\label{tab:mainresults}
\resizebox{0.8\linewidth}{!}{
\begin{tabular}{c|cc|ccc|cc|cc|cc}
\hline
\multirow{2}{*}{Method}	&\multicolumn{2}{c|}{TNL2K}
&\multicolumn{3}{c|}{AVisT}
&\multicolumn{2}{c|}{LaSOT}
&\multicolumn{2}{c|}{LaSOT$_{ext}$}
&\multicolumn{2}{c}{TrackingNet} \\
\cline{2-12}
&AUC	&P &AUC	&OP$_{0.5}$	&OP$_{0.75}$ &AUC	&P	&AUC	&P  &AUC	&P  \\ 
\hline
\multicolumn{12}{c}{\textbf{Performance-oriented Variants}}\\
\hline
\rowcolor{mygray}
UniSOT-L 		                        &\textbf{{62.8}}
                                        &\textbf{{64.6}}
                                        &\textbf{{57.8}}
                                        &\textbf{{67.9}}
                                        &\textbf{{48.7}}
                                        &\textbf{{71.3}}
                                        &\textbf{{78.3}}
                                        &\textbf{{51.2}}
                                        &\textbf{{59.0}}
                                        &\textbf{{84.1}}
                                        &82.9 \\
APMT-SwinB~\cite{apmt}                 &-&-&-&-&-
                                        &70.2&77.0 &-&-&83.6&82.2 \\
OSTrack-384~\cite{OSTrack}              &\underline{{55.9}}
                                        &-
                                        &-
                                        &-
                                        &-
                                        &\underline{{71.1}}
                                        &\underline{{77.6}}
                                        &\underline{{50.5}}
                                        &\underline{{57.6}}
                                        &\underline{{83.9}}
                                        &\textbf{{83.2}}\\
OmniTracker-L~\cite{Unicorn}    &-&-&-&-&-&69.1&75.4&-&-&83.4&82.3 \\
Unicorn-ConvL~\cite{Unicorn}    &-&-&-&-&-&68.5&74.1&-&-&83.0&82.2 \\
MixFormer-L~\cite{cui2022mixformer}     &-      &-    &\underline{{56.0}}   &\underline{{65.9}}   &\underline{{46.3}}  &70.1   &76.3   &-      &-      &\underline{{83.9}}     &\underline{{83.1}}       \\ 
SimTrack-L/14~\cite{SimTrack}           &55.6   &\underline{{55.7}}   &-      &-      &-    &70.5    &-      &-          &-      &83.4   &-         \\ 
STARK-ST101~\cite{yan2021learning}     &-&-&50.5&58.2&39.0&67.1&-&-&-&82.0&86.9 \\
\hline
\multicolumn{12}{c}{\textbf{Basic Variants}}\\
\hline
\rowcolor{mygray}
UniSOT-B 		                        &\textbf{{60.9}}
                                        &\textbf{{61.9}}
                                        &\textbf{{56.5}}
                                        &\textbf{{66.0}}
                                        &\textbf{{45.1}}
                                        &\underline{{69.4}}
                                        &{{74.9}}
                                        &\textbf{{49.2}}
                                        &\textbf{{55.8}}
                                        &\underline{{83.4}}
                                        &\textbf{{82.1}} \\
DiffusionTrack~\cite{diffusiontrack}    &56.5 &57.3 &-&-&- &\textbf{70.7}&\textbf{77.3}&-&-&\textbf{83.6}&\underline{82.0} \\
BIT~\cite{bit}                         &-&-&-&-&-&68.6&73.2&\underline{48.8}&\underline{56.8}&83.0&81.0\\
MCFT~\cite{mcft}                       &-&-&-&-&-&68.5&74.5&-&-&82.4&81.7\\
APMT-SwinT~\cite{apmt}                 &-&-&-&-&-
                                        &68.4&74.2&-&-&81.7&79.8\\
SFTransT~\cite{sftranst}               &54.6&\underline{55.5}&-&-&-&69.0&73.9&46.4&54.1&82.9&81.3 \\
OSTrack-256~\cite{OSTrack}              &54.3   &-      &-      &-      &-    &69.1  &\underline{{75.2}}   &47.4   &53.3   &{{83.1}}   &{{82.0}}       \\
MixFormer-22k~\cite{cui2022mixformer}       &-      &-      &\underline{{53.7}}   &\underline{{63.0}}   &\underline{{43.0}}   &69.2   &74.7   &-      &-      &{{83.1}}    &81.6       \\
SimTrack-B/16~\cite{SimTrack}           &\underline{{54.8}}   &53.8   &-      &-      &-    &{{69.3}}   &-      &-      &-      &82.3   &-          \\
AiATrack~\cite{AiATrack}                &-      &-      &-      &-      &-   &69.0   &73.8   &47.7     &55.4   &82.7   &80.4       \\
STARK~\cite{yan2021learning}		    &-      &-	    &51.1   &59.2   &39.1   &66.4	&71.2   &-      &-      &81.3	&78.1       \\
TransT~\cite{TransT}		            &50.7   &51.7   &49.0   &56.4   &37.2   &64.9	&73.8   &-      &-	    &81.4	&80.3       \\
TrDiMP~\cite{wang2021transformer}       &-      &-      &48.1   &55.3   &33.8   &63.9   &61.4   &-      &-      &78.4   &73.1       \\
Ocean~\cite{Ocean}		                &38.4   &37.7   &38.9   &43.6   &20.5   &56.0	&56.6	&-      &-      &-	    &-	        \\
KYS~\cite{KYS}	    	                &44.9   &43.5   &-&-&-&55.4	&-	    &-   &-      &74.0	&68.8	    \\
SiamFC++~\cite{SiamFC++}		        &38.6   &36.9   &-&-&-&54.4	&54.7   &-   &-  	&75.4	&70.5	    \\
SiamBAN~\cite{SiamBAN}		            &41.0   &41.7   &-&-&-&51.4	&52.1	&-   &-      &-	    &-	        \\
DiMP~\cite{DiMP}    	                &44.7   &43.4   &41.9   &45.7   &26.0   &56.9	&56.7	&39.2   &45.1   &74.0	&68.7       \\
SiamRPN++~\cite{SiamRPNplusplus}	    &41.3   &41.2   &39.0   &43.5   &21.2   &49.6	&49.1   &34.0   &39.6	&73.3	&69.4       \\
ECO~\cite{ECO}                          &32.6   &31.7   &-&-&- &32.4   &30.1   &22.0   &24.0   &55.4   &49.2       \\
MDNet~\cite{MDNet}                      &-      &-      &-&-&- &39.7   &37.3   &27.9   &31.8   &60.6   &56.5       \\
SiamFC~\cite{SiameseFC}                  &29.5   &28.6   &-      &-      &-      &33.6   &33.9   &23.0   &26.9   &57.1   &53.3       \\
\hline
\end{tabular}
}
\end{center}
\vspace{-4mm}
\end{table*}

\begin{table*}[t]
\begin{center}
\caption{Comparison with state-of-the-art trackers on NFS, UAV123 datasets in terms of overall AUC score. The best two results are shown in bold and underline.}
\label{tab:smallresults}
\resizebox{0.85\linewidth}{!}{
\begin{tabular}{c|cccccc|cc}
\hline
& Ocean
& DiMP-50
& PrDiMP-50
& TransT
& OSTrack-256 & OSTrack-384
& UniSOT-B  & UniSOT-L \\
\hline
NFS      &49.4    &61.8
                    &63.5    &65.3
                    &64.7    &\underline{{66.5}}
                    &65.9
                    &\textbf{{67.6}}\\
UAV123   &57.4    &64.3
                    & 68.0   &68.1
                    &68.3    &\underline{{70.7}}
                    &69.3
                    &\textbf{{71.0}}\\
\hline
\end{tabular}
}
\end{center}
\vspace{-4mm}
\end{table*}

\begin{table}[!t]
    \setlength{\tabcolsep}{2mm}
    \centering
    \caption{Comparison with state-of-the-art trackers on VOT2022.}
    \vspace{-2mm}
    \label{tab:vot2022}
    \begin{tabular}{c|ccc}
        \hline
        \multicolumn{1}{c|}{Method} &{EAO} $\uparrow$  &Acc $\uparrow$ &Rob $\uparrow$    \\
        \hline
        \multicolumn{4}{c}{Base Backbones} \\
        \hline  \rowcolor{mygray}
        UniSOT-B  & \textbf{0.558} & \textbf{{0.791}}  & \textbf{0.852} \\
        SeqTrack-B256~\cite{seqtrackv2}  & 0.513 & {0.789}  & 0.806 \\
        OSTrack-256~\cite{OSTrack}   & 0.516 & 0.777  & 0.801 \\
        MixFormer-22k~\cite{cui2022mixformer} & {0.531} & 0.775  & {0.842} \\
        SwinTrack-B~\cite{swintrack}   & 0.524 & 0.788  & 0.803 \\
        TransT~\cite{TransT}        & 0.512 & 0.781  & 0.800 \\
        STARK-ST50~\cite{yan2021learning}    & 0.467 & 0.778  & 0.756 \\
        DiMP~\cite{DiMP}          & 0.430 & 0.689  & 0.760 \\
        SiamFC~\cite{SiameseFC}        & 0.255 & 0.562  & 0.543 \\
        \hline
        \multicolumn{4}{c}{Larger Backbone} \\
        \hline \rowcolor{mygray}
        UniSOT-L  & \textbf{0.561} & {0.777}  & \textbf{0.855} \\
        SeqTrack-L256~\cite{seqtrackv2}  & 0.542 & 0.770 & 0.852 \\
        MixFormer-L~\cite{cui2022mixformer}    & 0.541 & \textbf{0.779} & 0.835 \\
        STARK-ST101~\cite{yan2021learning}     & 0.492 & 0.775  & 0.782 \\
        \hline
        \multicolumn{4}{c}{VOT2022 Challenge (unpublished or variants)} \\
        \hline
        DAMT~\cite{vot2022}        &0.602&0.776&0.887\\
        APMT-MR~\cite{vot2022}            & 0.591 & 0.787 & 0.877 \\
        OSTrackSTB~\cite{vot2022}  &0.591&0.790&0.869\\
        ADOTstb~\cite{vot2022}     &0.569&0.775&0.862\\
        SRATransT~\cite{vot2022}   &0.560&0.764&0.864\\
        \hline
    \end{tabular}
\end{table}
\begin{table}[!t]
    \centering
    \caption{Evaluation of UniSOT on VOTS2024.}
    \vspace{-2mm}
    \label{tab:vots2024}
    \begin{tabular}{c|ccc}
        \hline
        \multicolumn{1}{c|}{Method} &{Q} $\uparrow$  &Acc $\uparrow$ &Rob $\uparrow$    \\
        \hline  \rowcolor{mygray}
        UniSOT-L (+AlphaRefine) & \textbf{0.500} & 0.705  & \textbf{0.741} \\\rowcolor{mygray}
        UniSOT-B (+AlphaRefine)  & 0.481 & 0.699  & 0.716 \\
        MixFormer-L (+AlphaRefine)~\cite{vots2023} & 0.499 & \textbf{0.713} & 0.736 \\
        STARK (+AlphaRefine)~\cite{vots2023} & 0.462 & 0.695 & 0.685 \\
        MixFormerV2 (+AlphaRefine)~\cite{vots2023} & 0.458 & 0.678 & 0.693 \\
        ToMP (+AlphaRefine)~\cite{vots2023} & 0.442 & 0.681 & 0.671 \\
        \hline
        \multicolumn{4}{c}{Segmentation-based trackers}\\ \hline
        DMAOT~\cite{dmaot} & 0.636 & 0.751 & 0.795 \\
        Cutie+SAM~\cite{cutie} & 0.607 & 0.756 & 0.730 \\
        AOT~\cite{aot} & 0.550 & 0.698 & 0.767 \\
        \hline
    \end{tabular}
    \vspace{-3mm}
\end{table}

\begin{table}[!t]
\caption{Comparison with state-of-the-art vision-language trackers on LaSOT, TNL2K and OTB99 datasets.}
\label{tab:vlresults}
\centering
\begin{tabular}{c|cc|cc|cc}
\hline
\multirow{2}{*}{Method}
&\multicolumn{2}{c|}{TNL2K}
&\multicolumn{2}{c|}{LaSOT}
&\multicolumn{2}{c }{OTB99} \\
\cline{2-7}
&AUC   &P   &AUC   &P   &AUC   &P   \\
\hline
\multicolumn{7}{c}{\textbf{NL}}\\
\hline
\rowcolor{mygray}
UniSOT-L                      &\textbf{{56.4}}     &\textbf{{57.5}}    &\textbf{{59.6}}     &\textbf{{63.9}}          &\textbf{{63.5}}    &\textbf{{83.2}}        \\
\rowcolor{mygray}
UniSOT-B                      &\underline{{53.9}}    &\underline{{54.4}}    &\underline{{57.2}}    &\underline{{61.0}}        &60.1   &79.1       \\
QueryNLT        &53.3   &53.0    &54.2  &55.0  &\underline{61.2}  & \underline{81.0} \\
JointNLT	    &52.4    &52.2     &56.9    &59.3        &59.2   &77.6	    \\
CTRNLT    	        &14.0    &9.0      &52.0    &51.0        &53.0   &72.0	    \\
TNL2K-1		    &11.0    &6.0      &51.0    &49.0        &19.0   &24.0       \\
GTI                  &-       &-        &47.8    &47.6        &58.1   &73.2	    \\
RVTNLN      &-&-&-&-&54.0   &56.0\\
RTTNLD      &-&-&-&-&54.0   &78.0\\
TNLS-II          &-       &-        &-       &-           &25.0   &29.0       \\
\hline
\multicolumn{7}{c}{\textbf{NL+BBOX}}\\
\hline
\rowcolor{mygray}
UniSOT-L                      &\textbf{{63.0}}     &\textbf{{65.0}}       &\textbf{{71.4}}     &\textbf{{78.7}}          &\underline{{71.1}}    &\underline{{92.0}}        \\
\rowcolor{mygray}
UniSOT-B                      &\underline{{61.2}}    &\underline{{62.8}}        &\underline{{69.4}}    &\underline{{75.9}}        &69.3   &89.9       \\
QueryNLT        &57.8   &58.7   &59.9    &63.5  &66.7   &88.2 \\
JointNLT	    &54.6    &54.8    &60.4    &63.6        &65.3   &85.6	    \\
VLTTT    	        &53.1    &53.3    &67.3    &72.1        &\textbf{{76.4}}   &\textbf{{93.1}}	    \\
SNLT               &27.6    &41.9    &54.0    &57.6        &66.6   &80.4	    \\
TNL2K-2		    &42.0    &42.0    &51.0    &55.0        &68.0   &88.0       \\
RVTNLN      &25.0   &27.0   &50.0   &56.0   &67.0  &73.0 \\
RTTNLD      &25.0   &27.0   &35.0   &35.0   &61.0  &79.0 \\
TNLS-III         &-       &-       &-       &-           &55.0   &72.0       \\
\hline
\end{tabular}
\end{table}

\begin{table*}[!t]
\begin{center}
\caption{Comparison with state-of-the-art grounding methods on RefCOCO, RefCOCO+, RefCOCOg datasets. $^\dagger$ means UniSOT is trained with the general visual grounding setting}
\vspace{-2mm}
\label{tab:grresults}
\resizebox{0.7\linewidth}{!}{
\begin{tabular}{c|ccc|ccc|ccc}
\hline
\multirow{2}{*}{Method}
&\multicolumn{3}{c|}{RefCOCO}
&\multicolumn{3}{c|}{RefCOCO+}
&\multicolumn{3}{c }{RefCOCOg} \\
\cline{2-10}
&\textit{val} &\textit{testA} &\textit{testB} &\textit{val} &\textit{testA} &\textit{testB} &\textit{val-g} &\textit{val-u} &\textit{test-u}   \\
\hline
\rowcolor{mygray}
UniSOT$^\dagger$                        &\textbf{{85.47}}     &\textbf{{87.56}}     &\underline{{81.73}}     &\textbf{{74.60}}     &\textbf{{79.70}}    &\textbf{{65.64}}    &\textbf{{73.86}}    &\underline{{75.94}}    &\textbf{{74.86}}        \\
VLTVG  &\underline{{84.77}}   &\underline{{87.24}}   &80.49   &\underline{{74.19}}   &\underline{{78.93}}  &\underline{{65.17}}  &\underline{{72.98}}  &\textbf{{76.04}}    &74.18      \\
SeqTR       &83.72   &86.51   &81.24   &71.45   &76.26  &64.88  &71.50        
&74.86    &\underline{{74.21}}      \\
QRNet     &84.01   &85.85   &\textbf{{82.34}}   &72.94   &76.17  &63.81  &71.89
&73.03    &72.52      \\
TransVG  &81.02   &82.72   &78.35   &64.82   &70.70  &56.94  &67.02  &68.67    &67.73      \\
Ref-NMS      &80.70   &84.00   &76.04   &68.25   &73.68  &59.42  &-
&70.55    &70.62                      \\
LBYL-Net    &79.67 &82.91  &74.15   &68.64   &73.38  &59.49  &62.70
&-    &-      \\
ReSC-Large &77.63 &80.45  &72.30   &63.59   &68.36  &56.81  &63.12
&67.30    &67.20\\
NMTree   &76.41   &81.21   &70.09   &66.46   &72.02  &57.52  &64.62  
&65.87    &66.44      \\
\hline
\end{tabular}
}
\end{center}
\vspace{-5mm}
\end{table*}

\noindent
\textbf{Training Settings.}
As shown in Figure~\ref{fig:train}, two search regions interact with target references separately in the reference-generalized feature extractor.
Then, these embeddings of two search regions serve as video contexts for each other to provide scenario cues in the reference-adaptive box head.
Finally, the outputs of both search regions are constrained by the designed losses.
The loss weights are set to $\lambda_{giou}=2.0$, $\lambda_1=5.0$, $\lambda_{mmc}=0.1$.
In addition, our model is trained with a batch size of 64 and iterates 300 epochs with $3\times10^4$ samples per epoch.
AdamW optimizer~\cite{AdamW} with weight decay $10^{-4}$ is applied for model optimization.
The learning rates start from $4\times10^{-5}$ for the reference-generalized feature extractor and $4\times10^{-4}$ for the reference-adaptive box head, then gradually reduce to 0 with cosine annealing~\cite{CosLR}.
\subsubsection{The Second Training Stage.}
\label{sec:4.1.3}
\noindent
\textbf{Data Unit.}
We sample one template and two search regions as the first training stage does.
Further, we also sample different auxiliary video modalities corresponding to the template and search regions for video modality-joint tuning.
The inputs of auxiliary video modalities are converted to three channel formats as ViPT~\cite{ViPT} does.
The RGB+Depth, RGB+Thermal, RGB+Event data are sampled from DepthTrack~\cite{depthtrack}, LasHeR~\cite{lasher} and VisEvent~\cite{visevent} respectively, which is same as UnTrack~\cite{UnTrack}.

\noindent
\textbf{Fine-tuning Settings.}
We inject auxiliary modality tuning blocks (AMTB) into key, query, value projection of attention mechanisms and multi-layer perceptron.
The inherent rank of incremental weights $r$ is set as 32.
The block-average modality-shared rank budget $\hat{n}$ is set as 16.
The block-average modality-specific rank budget $\hat{m}$ is set as 32.
Thus, given the number of AMTB $N_{AMTB}$, modality-shared rank budget $n=\hat{n}\times N_{AMTB}$ and modality-specific rank budget $m=\hat{m}\times N_{AMTB}$.
We freeze the parameters of UniSOT trained in the first stage, and fine-tune incremental weights in AMTB with a batch size of 32 and iterate 60 epochs with $3\times10^4$ samples per epoch.
$\lambda_{orth}$ is set as 0.1.
AdamW optimizer~\cite{AdamW} with weight decay $10^{-4}$ is applied for model optimization.
The learning rates start from $8\times10^{-4}$ and then gradually reduce to 0 with cosine annealing.
\subsubsection{Inference Details}
\textbf{Efficient Inference.} We can remove unavailable references from inputs and replace all task-oriented multi-head attention with vanilla multi-head attention for efficient inference. 
In AMTB, we remove parameters in $P$ and $Q$ whose corresponding singular value $\Lambda$ is zero to speed up inference for RGB+X inputs.

\noindent
\textbf{Scenario Token Update.} 
Search region embeddings with high-confidence bounding boxes are saved as video contexts $\mathbf{E}_c^{N+M}$ to help with subsequent target localization.
Specifically, if the target confidence score ${\rm max}[\hat{\mathbf{C}}(x,y)\hat{\mathbf{L}}(x,y)]$ is higher than 0.5, the corresponding search region embeddings will be saved as video contexts.
The target token, distractor token and background token in the reference-adaptive box head are updated every 20 frames.

\subsection{State-of-the-art Comparisons}
%
\noindent
\textbf{Visual Tracking.}
We evaluate our trackers on seven visual tracking benchmarks, including TNL2K~\cite{TNL2K}, AVisT~\cite{AVisT}, LaSOT~\cite{LaSOT}, LaSOT$_{ext}$~\cite{LaSOT_ext}, TrackingNet~\cite{trackingnet}, NFS~\cite{NFS} and UAV123~\cite{UAV}, which are commonly used for visual tracker evaluation.
The Area Under the Curve (AUC) of the success plot is the main metric for ranking trackers.
%
As shown in Table~\ref{tab:mainresults}-\ref{tab:vots2024},
our UniSOT-B and UniSOT-L outperform trackers specifically designed for the visual tracking task on most benchmarks. 
%
Notably, UniSOT supports different reference modalities.
%
%
These results demonstrate the effectiveness of UniSOT using bounding box reference (\textbf{BBOX}).
We evaluate recent vision-language trackers initialized by the target bounding box on LaSOT.
JointNLT achieves 54.5\% AUC and VLTTT achieves 53.4\% AUC.
These vision-language trackers show limited performance without natural language.
This is because these trackers ignore the semantic gap between different reference modalities and tend to rely on semantic information in language.
Differently, UniSOT aligns vision and language into a unified semantic space, providing a unified tracking framework, which delivers superior performance for all three reference modalities.


\noindent
\textbf{Vision-Language Tracking.}
We further evaluate our UniSOT on vision-language tracking benchmarks, including TNL2K~\cite{TNL2K}, LaSOT~\cite{LaSOT} and OTB99~\cite{nltrack} and compare with the latest trackers, including QueryNLT~\cite{querynlt}, JointNLT~\cite{JointNLT}, CTRNLT~\cite{TSN}, TNL2K~\cite{TNL2K}, GTI~\cite{GTI}, TNLS~\cite{nltrack}, VLTTT~\cite{VLT}, RVTNLN~\cite{RVTNLN}, RTTNLD~\cite{rttnld}, SNLT~\cite{SNLT}.
As shown in Table~\ref{tab:vlresults}, when specifying target object by natural language (\textbf{NL}), UniSOT-L surpasses the previous best tracker JointNLT with a large margin on three benchmarks.
When initializing the tracker with both natural language and the bounding box (\textbf{NL+BBOX}), our UniSOT-L achieves the best performance on TNL2K and LaSOT.
These results demonstrate the superiority of UniSOT for vision-language tracking.

\noindent
\textbf{Efficiency.}
UniSOT-B runs at 58 FPS for visual tracking and 57 FPS for vision-language tracking.
UniSOT-L runs at 28 FPS for visual tracking and 27 FPS for vision-language tracking.
Compared with JointNLT (39 FPS), UniSOT-B achieves better performance with \textbf{1.46$\times$} speed.

\noindent
\textbf{Visual Grounding.}
Like VLTVG, we retrain UniSOT-B on train sets of RefCOCO~\cite{RefCOCO}, RefCOCO+~\cite{RefCOCO} and RefCOCOg~\cite{RefCOCOg} separately for a fair comparison, and report the Top-1 accuracy on corresponding test sets in Table~\ref{tab:grresults}.
We compared UniSOT$^\dagger$ with the grounding methods, including VLTVG~\cite{yang2022improving}, SeqTR~\cite{zhu2022seqtr}, QRNet~\cite{ye2022shifting}, TransVG~\cite{deng2021transvg}, Ref-NMS~\cite{chen2021ref}, LBYL-Net~\cite{huang2021look}, ReSC-Large~\cite{yang2020improving}, NMTree~\cite{liu2019learning}.
Here, UniSOT$^\dagger$ means UniSOT is retrained with the general visual grounding setting~\cite{yang2022improving}.
The test image is scaled such that its long edge is 384.
Although UniSOT is not specifically designed for grounding, UniSOT achieves the best performance on seven test sets, demonstrating the generalization of our framework.

\noindent
\textbf{Discussion.}
As mentioned above, existing trackers are designed for single or partial reference modalities and overspecialize on the specific reference modality.
UniSOT not only can cope with three types of target reference (NL, BBOX, NL+BBOX) but also outperforms previous reference modality-specific counterparts on 12 visual and vision-language tracking benchmarks, which proves the effectiveness and generalization of UniSOT.

\begin{table}[t]
  \centering
  \setlength{\tabcolsep}{2mm}{
    \small
        \caption{Comparison with state-of-the-art RGBD trackers. * means only train with the first stage, similarly hereinafter.}
        \vspace{-2mm}
        \label{tab:rgbd}
        \resizebox{1.0\linewidth}{!}{
            \begin{tabular}{c|ccc|ccc}
                \hline
                \multirow{2}*{Method} & \multicolumn{3}{c}{VOT-RGBD22} & \multicolumn{3}{c}{DepthTrack} \\
                    \cline{2-7}
                & EAO & Acc. & Rob. &F-score &Re &Pr \\
                \hline
                \multicolumn{7}{c}{Unified Model with Uniform Parameters} \\
                \hline
                \rowcolor{mygray} UniSOT-L &\textbf{73.8}&\textbf{82.1}&\textbf{89.1}&\textbf{63.2}&\textbf{62.5}&\textbf{63.9}\\
                \rowcolor{mygray} UniSOT-B &\underline{73.2}&{81.8}&\underline{88.8}&\underline{62.5}&\underline{62.2}&\underline{62.9}\\
                \rowcolor{mygray} UniSOT-L* & 68.9 & 81.6& 84.3& 59.4& 60.6 & 58.3\\
                \rowcolor{mygray} UniSOT-B* & 66.6& 80.2& 81.3& 53.8& 54.9& 52.7\\
                Un-Track~\cite{UnTrack} &71.8&\underline{82.0} &86.4 &61.0&61.0&61.0\\
                \hline
                \multicolumn{7}{c}{Unified Model with Separate Parameters} \\
                \hline
                ViPT~\cite{ViPT} &72.1&81.5 &87.1 &59.4&59.6&59.2\\
                ProTrack~\cite{protrack} &65.1&80.1&80.2 &57.8&57.3&58.3\\
                \hline
                \multicolumn{7}{c}{Specialized Tracker for RGBD Tracking or VOT Challenge} \\
                \hline
                MixForRGBD~\cite{vot2022} &77.9&81.6&94.6 &-&-&-\\
                SAMF~\cite{vot2022} &76.2&80.7&93.6 &-&-&-\\
                ProMix~\cite{vot2022} &72.2&79.8&90.0 &-&-&-\\
                SBT-RGBD~\cite{vot2022} &70.8&80.9&86.4 &-&-&-\\
                SPT~\cite{rgbd1k} &65.1&79.8&85.1 &53.8&54.9&52.7\\
                DMTrack~\cite{vot2022} &65.8&75.8&85.1 &-&-&-\\
                DeT~\cite{depthtrack} &65.7&76.0&84.5 &53.2&50.6&56.0\\
                DDiMP~\cite{vot2020} &-&-&- &48.5&56.9&50.3\\
                STARK-RGBD~\cite{yan2021learning} &64.7&80.3&79.8 &-&-&-\\
                DRefine~\cite{vot2021} &59.2&77.5&76.0 &-&-&-\\
                ATCAIS~\cite{vot2020} &55.9&76.1&73.9 &47.6&45.5&50.0\\
                \hline
                \multicolumn{7}{c}{Non-real Time Trackers} \\
                \hline
                SeqTrackv2-L384~\cite{seqtrackv2} &74.8&82.6&91.0 &62.3&62.6&62.5\\
                SeqTrackv2-L256~\cite{seqtrackv2} &74.9&81.3&91.8 &62.8&63.0&62.5\\
                \hline
            \end{tabular}   
        }
    }
\end{table}

\begin{table}[t]
  \centering
  \setlength{\tabcolsep}{3mm}{
    \small
        \caption{Comparison with state-of-the-art RGBT trackers.}
        \vspace{-2mm}
        \label{tab:rgbt}
        \resizebox{0.82\linewidth}{!}{
            \begin{tabular}{c|cc|cc}
            \hline
            \multirow{2}*{Method} & \multicolumn{2}{c}{LasHeR} & \multicolumn{2}{c}{RGBT234} \\
                \cline{2-5}
            & AUC & P &MSR &MPR \\
            \hline
            \multicolumn{5}{c}{Unified Model with Uniform Parameters} \\
            \hline
            \rowcolor{mygray} UniSOT-L &\textbf{54.8}&\textbf{68.4} &\textbf{65.0}&\textbf{87.4}\\
            \rowcolor{mygray} UniSOT-B &\underline{54.0}&\underline{67.4} &\underline{64.1}&\underline{86.6}\\
            \rowcolor{mygray} UniSOT-L* & 43.4& 55.9& 63.3& 84.2\\
            \rowcolor{mygray} UniSOT-B* & 42.9& 55.5& 61.6& 82.2\\
            Un-Track~\cite{UnTrack} &51.3&64.6 &62.5&84.2\\
            \hline
            \multicolumn{5}{c}{Unified Model with Separate Parameters} \\
            \hline
            ViPT~\cite{ViPT} &52.5&65.1 &61.7&83.5\\
            ProTrack~\cite{protrack} &42.0&53.8 &59.9&79.5\\
            \hline
            \multicolumn{5}{c}{Specialized Tracker for RGBT Tracking} \\
            \hline
            M3PT~\cite{m3pt} &54.2 &67.3 &63.4&85.9 \\
            CMD~\cite{cmd} &46.6 &59.0 &58.4 & 82.4 \\
            APFNet~\cite{rgbt2} &36.2&50.0 &57.9&82.7\\
            CMPP~\cite{cmpp} &-&- &57.5&82.3\\
            JMMAC~\cite{jmmac} &-&- &57.3&79.0\\
            mfDiMP~\cite{mfdimp} &34.3&44.7 &42.8&64.6\\
            DAPNet~\cite{dapnet} &31.4&43.1 &-&-\\
            CAT~\cite{cat} &31.4&45.0 &56.1&80.4\\
            HMFT~\cite{vtuav} &31.3&43.6 &-&-\\
            MaCNet~\cite{macnet} &-&- &55.4&79.0\\
            FANet~\cite{fanet} &30.9&44.1 &55.3&78.7\\
            DAFNet~\cite{dafnet} &-&- &54.4&79.6 \\
            SGT~\cite{sgt} &25.1&36.5 &47.2&72.0 \\
            \hline
            \multicolumn{5}{c}{Non-real Time Trackers} \\
            \hline
            SeqTrackv2-L384~\cite{seqtrackv2} & 61.0 & 76.7 & 68.0 & 91.3 \\
            SeqTrackv2-L256~\cite{seqtrackv2} & 58.8 & 74.1 & 68.5 & 92.3 \\
            \hline
            \end{tabular}  
        }
    }
\end{table}

\noindent
\textbf{RGB-Depth Tracking.}
VOT-RGBD22~\cite{vot2022} is the most recent dataset in RGB-D tracking, consisting of 127 short-term RGB-D sequences.
DepthTrack~\cite{depthtrack} is a comprehensive long-term RGBD tracking benchmark, which contains 150 training and 50 testing videos featuring 15 per-frame attributes.
We evaluate UniSOT on DepthTrack and VOT-RGBD22 and compare it with the state-of-the-art RGBD trackers in Table~\ref{tab:rgbd}.
UniSOT-B achieves 73.2\% EAO on VOT-RGBD22 and 62.5\% F-score on DepthTrack, surpassing Un-Track with a large margin.
With a stronger foundation tracker, UniSOT-L achieves higher performance on both VOT-RGBD22 and DepthTrack.

\noindent
\textbf{RGB-Thermal Tracking.}
LasHeR~\cite{lasher} is a large-scale dataset with high diversity for RGB+Thermal tracking, which comprises 979 videos for training and 245 videos for testing.
RGBT234~\cite{rgbt234} is a large-scale RGBT tracking dataset containing 234 videos and adopts MSR and MPR as evaluation criteria.
As shown in Table~\ref{tab:rgbt}, UniSOT-B outperforms Un-Track trackers with 2.7\% AUC on LasHeR and 2.5\% MSR on RGBT234.
Meanwhile, UniSOT-L obtains strong performance of 54.8\% AUC on LasHeR and 65.0\% MSR on RGBT234 surpassing Un-Track with a large margin.

\noindent
\textbf{RGB-Event Tracking.}
VisEvent~\cite{visevent} is the largest benchmark for RGBE tracking, which contains 500 videos for training and 320 videos for testing.
Table~\ref{tab:rgbe} shows that UniSOT-B achieves 60.7\% AUC, which surpasses the Un-Track with 1.8\% AUC.
Furthermore, UniSOT-L achieves the AUC score of 62.0\%, which performs best among real-time trackers.

\noindent
\textbf{Discussion.}
UniSOT-B and UniSOT-L are running at 48 FPS and 21 FPS with auxiliary video modalities.
The notation UniSOT-L* and UniSOT-B* in Table~\ref{tab:rgbd}-\ref{tab:rgbe} indicates that these models are trained only with RGB data in the first training stage.
We can find that UniSOT only achieves limited performance without the second training stage.
This is because RGB images have inherent shortcomings in challenging scenarios that have few appearance cues.
After two stages of training, UniSOT achieves significantly robust performance with auxiliary video modalities, reflecting their necessity in complex scenarios.
Further, our UniSOT achieves competitive performance among video modality-specific counterparts on all three auxiliary video modalities (RGBD, RGBT, RGBE).
Notably, UniSOT only needs to be fine-tuned once to achieve high-performance tracking with all three auxiliary video modalities.
Meanwhile, different video modalities share a unified parameter set, which avoids redundant video modality-specific parameter sets.
Recently, Un-Track designs a unified architecture and learns video modality-specific features by projecting auxiliary video modalities into different low-rank spaces with an identical rank.
Such a design poses a challenge for video modality-aligned feature learning and ignores the potential difference in the amount of information across video modalities, which may desire various ranks.
Conversely, UniSOT dynamically adjusts the rank of shared parameters for both video modality-aligned and video modality-specific feature learning, enabling UniSOT to exceed Un-Track by a large margin.
Moreover, UniSOT supports tracking with different reference modalities (NL, BBOX, NL+BBOX) in sequences of different video modalities (RGB, RGB+X) under a unified architecture with uniform parameters, which has more generalized application scenarios.
More combinations of reference and video modalities are discussed in Supplementary Materials.

\begin{table}[t]
    \caption{Comparison with state-of-the-art RGBE trackers.}
    \vspace{-2mm}
    \label{tab:rgbe}
    \centering
    \setlength{\tabcolsep}{6mm}{
        \small
        \resizebox{0.8\linewidth}{!}{
        \begin{tabular}{c|cc}
        \hline
        \multirow{2}*{Method} & \multicolumn{2}{c}{VisEvent}\\
        \cline{2-3}
        & AUC & P \\
        \hline
        \multicolumn{3}{c}{Unified Model with Uniform Parameters} \\
        \hline
        \rowcolor{mygray} UniSOT-L &\textbf{62.0}&\textbf{79.0} \\
        \rowcolor{mygray} UniSOT-B &\underline{60.7}&\underline{78.0} \\
        \rowcolor{mygray} UniSOT-L* & 53.2& 69.2\\
        \rowcolor{mygray} UniSOT-B* & 52.6& 69.6\\
        Un-Track~\cite{UnTrack} &58.9 &75.5\\
        \hline
        \multicolumn{3}{c}{Unified Model with Separate Parameters} \\
        \hline
        ViPT~\cite{ViPT} &59.2&75.8\\
        ProTrack~\cite{protrack} &47.1&63.2\\
        \hline
        \multicolumn{3}{c}{Specialized Tracker for RGBE Tracking} \\
        \hline
        TransT\_E~\cite{TransT} &47.4&65.0\\
        PrDiMP\_E~\cite{PrDiMP} &45.3&64.4 \\
        STARK\_E~\cite{yan2021learning} &44.6&61.2 \\
        MDNet\_E~\cite{MDNet} &42.6&66.1 \\
        VITAL\_E~\cite{VITAL} &41.5&64.9 \\
        ATOM\_E~\cite{ATOM} &41.2&60.8 \\
        SiamBAN\_E~\cite{SiamBAN} &40.5&59.1 \\
        SiamMask\_E~\cite{SiamMask} &36.9&56.2 \\
        \hline
        \multicolumn{3}{c}{Non-real Time Trackers} \\
        \hline
        SiamRCNN\_E~\cite{SiamRCNN} &49.9&65.9 \\
        SeqTrackv2-L384~\cite{seqtrackv2} & 63.4 & 80.0 \\
        SeqTrackv2-L256~\cite{seqtrackv2} & 63.0 & 79.4 \\
        \hline
        \end{tabular}
        }
    }
    \vspace{-3mm}
\end{table}

\subsection{Ablation Study} \label{sec:4.3}
The following experiments use UniSOT-B as the base model.
The baseline is UniSOT without MMCLoss constraint and using the anchor-free head for localization.
FPS is test under vision-language tracking.

\begin{table}[t]
\caption{Analysis of different components in UniSOT.}
\label{tab:components}
\centering
\begin{tabular}{c|cc|cc|cc|c}
\hline
\multirow{3}{*}{Method}
&\multicolumn{6}{c|}{TNL2K} & \multirow{3}{*}{FPS}\\
\cline{2-7}
&\multicolumn{2}{c|}{BBOX}
&\multicolumn{2}{c|}{NL}
&\multicolumn{2}{c|}{NL+BBOX} \\
\cline{2-7}
&AUC &P &AUC &P &AUC &P  & \\
\hline
baseline            &57.3     &57.8     &49.6    &49.8    &57.4   &58.5  & 58 \\
+MMCLoss            &58.7    &60.2    &51.9    &52.3    &59.8   &61.2 & 58  \\
\rowcolor{mygray}
+RABH            &\textbf{60.9}     &\textbf{61.9}     &\textbf{53.9}     &\textbf{54.4}     &\textbf{61.2}    &\textbf{62.8}  & 57  \\
\hline
\end{tabular}
\vspace{-4mm}
\end{table}

\begin{figure}[t]
    \centering
    \includegraphics[width=1.0\linewidth]{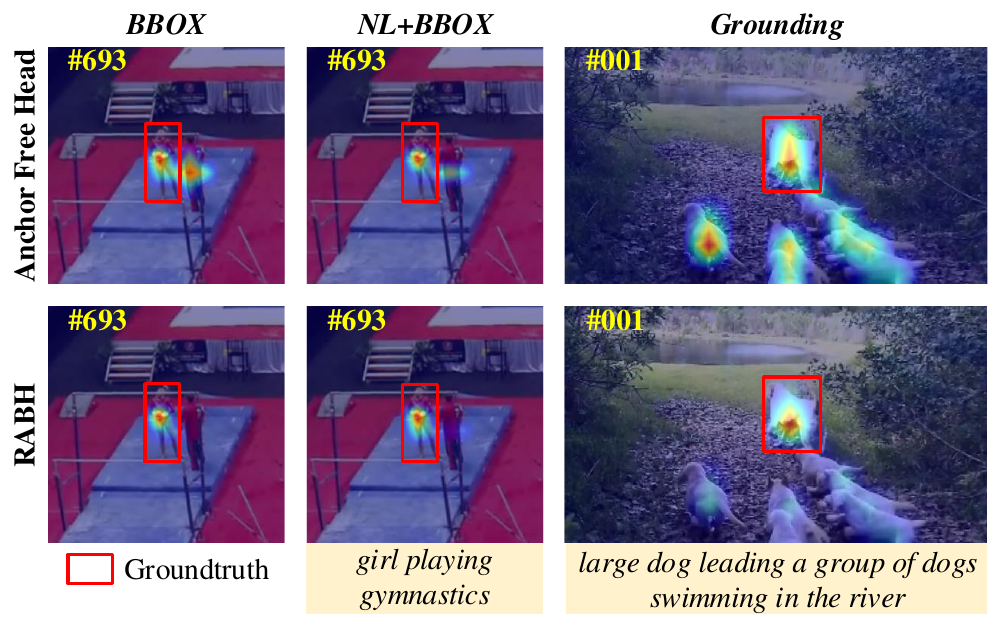}
    \caption{
    Visualization of target localization results. RABH obtains more stable localization results. This is because RABH can adaptively capture scenario cues and localize target object in a contrastive way, thus suppressing distractors.
    }
    \label{fig:visualization}
    \vspace{-3mm}
\end{figure}

\begin{table}[t]
\caption{Analysis of the reference-generalized feature extractor.}
\label{tab:backbone}
\centering
\begin{tabular}{c|c|cc|cc|cc|c}
\hline
\multirow{3}{*}{$N$}
&\multirow{3}{*}{$M$}
&\multicolumn{6}{c|}{TNL2K} & \multirow{3}{*}{FPS}\\
\cline{3-8}
&&\multicolumn{2}{c|}{BBOX}
&\multicolumn{2}{c|}{NL}
&\multicolumn{2}{c|}{NL+BBOX} \\
\cline{3-8}
&&AUC &P &AUC &P &AUC &P   \\
\hline
0 & 12 &59.7     &60.6     &53.4     &53.6     &60.4    &61.2  & 59  \\
3 & 9  &59.9     &60.8     &\textbf{54.1}     &\textbf{54.5}     &61.1    &62.7  &58   \\
\rowcolor{mygray}
6 & 6  &\textbf{60.9}     &\textbf{61.9}     &{53.9}     &{54.4}     &\textbf{61.2}    &\textbf{62.8}  & 57   \\
9 & 3  &60.8     &61.8     &53.0     &52.9     &60.9    &62.0   &55  \\
11 & 1 &60.7     &61.6     &45.1     &41.8     &60.3    &61.3   &54  \\
\hline
\end{tabular}
\end{table}

\begin{table}[!t]
\caption{Analysis of the multi-modal contrastive loss designing.}
\label{tab:mmcloss}
\centering
\setlength{\tabcolsep}{1.5mm}{
\begin{tabular}{c|c|c|cc|cc|cc}
\hline
\makebox[11pt][c]{\multirow{3}{*}{pos.}}
&\multirow{3}{*}{neg.}
&\makebox[14pt][c]{\multirow{3}{*}{$N_{neg}$}}
&\multicolumn{6}{c}{TNL2K}\\
\cline{4-9}
&&&\multicolumn{2}{c|}{BBOX}
&\multicolumn{2}{c|}{NL}
&\multicolumn{2}{c}{NL+BBOX} \\
\cline{4-9}
&&&\makebox[11pt][c]{AUC} &P &\makebox[11pt][c]{AUC} &P &\makebox[11pt][c]{AUC} &P   \\
\hline
\makebox[11pt][c]{$avg$} & \makebox[15pt][c]{$rand$} & 9  &59.4    &60.2     &51.8     &51.1     &59.5    &60.2     \\
$ctr$ & \makebox[15pt][c]{$rand$} & 9  &59.9    &60.8     &53.0     &53.1     &60.1    &61.3     \\
\makebox[11pt][c]{$avg$} & $top$  & 9  &60.3    &61.1     &53.5     &53.5     &60.4    &61.7     \\
\rowcolor{mygray}
$ctr$ & $top$  & 9  &\textbf{60.9}     &\textbf{61.9}     &\textbf{53.9}     &\textbf{54.4}     &\textbf{61.2}    &\textbf{62.8}     \\
$ctr$ & $top$  & 1  &59.7&60.6&52.5&52.6&59.9&61.1     \\
$ctr$ & $top$  & 5  &60.4&61.2&53.5&53.8&60.7&62.1     \\
$ctr$ & $top$  & 13 &60.8&61.8&53.8&54.2&\textbf{61.2}&62.7     \\
\hline
\end{tabular}
}
\vspace{-3mm}
\end{table}

\noindent
\textbf{Effectiveness of the Different Components.}
Table~\ref{tab:components} shows the performance of UniSOT with different components.
Due to the semantic gap between vision and language, different reference modalities may push the baseline model toward divergent optimization directions during joint training, resulting in compromised performance for all reference modalities.
MMCLoss brings 1.4\%, 2.3\% and 2.4\% AUC gains for BBOX, NL and BBOX+NL reference modalities respectively.
This is because MMCLoss can align different modal features into a unified semantic space, which enables consistent feature learning for different reference modalities.
The reference-adaptive box head (RABH) brings 2.2\%, 2.0\% and 1.4\% AUC gains for BBOX, NL and BBOX+NL reference modalities respectively with minimal efficiency sacrifice.
As shown in Figure~\ref{fig:visualization}, the anchor free head shows limited target localization results.
The reason is that enhanced features of search region vary with reference modalities, while the box head is fixed in tracking, which leads to compromising results across reference modalities.
Differently, we design a dynamic head (RABH), which can make full use of reference information to localize target object in a contrastive way, improving tracking performance across all reference modalities.

\noindent
\textbf{Analysis of the Reference-Generalized Feature Extractor.}
%
As shown in Table~\ref{tab:backbone}, more separate layers (larger $N$) are beneficial for visual tracking and more fusion layers (larger $M$) are beneficial for vision-language tracking.
Moreover, the efficiency is reduced by more separate layers as the serial inference of vision and language encoders.
However, when we fuse visual and language features in all encoder layers, the performance is suboptimal for vision-language tracking.
%
This is because early fusion breaks the low-level feature modeling for different reference modalities.
Thus, we set $N$=6 and $M$=6 to balance the performance for all reference modalities and efficiency.

\noindent
\textbf{Analysis of the Multi-Modal Contrastive Loss.}
We study different ways to obtain the positive sample and the negative sample.
As shown in Table~\ref{tab:mmcloss}, the best results are achieved when we sample the central score of target object as the positive sample and the top 9 scores out of the target box as negative samples.
The underlying reason is that the central feature of target object contains no backgrounds, which is more reliable for expressing target object.
Further, more hard-negative samples can improve the discriminability of the semantic token, while an excess of such samples may introduce easy-to-distinguish samples, thereby diluting the weight of difficult-to-distinguish (distractor) samples in MMCLoss and weakening the discriminability optimization of the semantic token toward distractors.
%
%

\noindent
\textbf{Analysis of the Distractor Threshold.}
As shown in Table~\ref{tab:distractor_thres}, the performance of UniSOT is insensitive to threshold $\beta$ over a wide range.
However, the performance drops overtly if we aggregate all background features into one token ($\beta$=0).
This is because the distractor feature is vital to discriminate target object in complex scenarios.
If we aggregate all background features together, the distractor features will be smoothed by other background features, which is not conducive for the tracker to distinguish distractors.

\begin{table}[!t]
\caption{Analysis of the distractor threshold $\beta$ in the distribution-based cross-attention mechanism.}
\label{tab:distractor_thres}
\centering
\begin{tabular}{c|cc|cc|cc}
\hline
\multirow{3}{*}{$\beta$}
&\multicolumn{6}{c}{TNL2K}\\
\cline{2-7}
&\multicolumn{2}{c|}{BBOX}
&\multicolumn{2}{c|}{NL}
&\multicolumn{2}{c}{NL+BBOX} \\
\cline{2-7}
&AUC &P &AUC &P &AUC &P   \\
\hline
0.00               &59.5&60.1&52.5&52.8&60.0&60.7     \\
0.25               &60.3&61.1&53.2&53.5&60.5&61.7     \\
0.50               &60.6&61.6&53.6&54.1&60.9&62.5     \\
\rowcolor{mygray}
0.75               &\textbf{60.9}     &\textbf{61.9}     &\textbf{53.9}     &\textbf{54.4}     &\textbf{61.2}    &\textbf{62.8}     \\
0.85               &60.8&61.8&53.6&53.8&60.9&62.4     \\
\hline
\end{tabular}
\end{table}

\begin{table}[t]
\caption{Analysis of the Robustness under Extreme Scenarios.}
\label{tab:noisy_datsaet}
\centering
\begin{tabular}{c|cc|cc|cc}
\hline
\multirow{3}{*}{Method}
&\multicolumn{6}{c}{OTB99}\\
\cline{2-7}
&\multicolumn{2}{c|}{BBOX}
&\multicolumn{2}{c|}{NL}
&\multicolumn{2}{c}{NL+BBOX} \\
\cline{2-7}
&AUC &P &AUC &P &AUC &P  \\
\hline
Original Image      &{69.0}     &{89.4}     &{60.1}     &{79.1}     &{69.3}    &{89.9}  \\
Low-contrast Image &68.3     &88.5     &59.6   &78.6    &68.7    &89.2   \\
Noisy Image      &68.1     &88.2     &59.2   &78.4    &68.5    &88.8   \\
\hline
\end{tabular}
\vspace{-4mm}
\end{table}

\noindent
\textbf{Analysis of the Robustness under Extreme Scenarios.}
To evaluate the performance of UniSOT  under sparse or noisy conditions, we constructed sparse and noisy datasets by reducing contrast and adding noise on OTB99 dataset. 
Experimental results in Table~\ref{tab:noisy_datsaet} demonstrate that UniSOT maintains comparable performance in both sparse and noisy datasets.
These evaluations validate the robustness and generalization capability of our framework across diverse and extreme conditions.

\begin{table}[!t]
\caption{Analysis of the model auxiliary modality tuning block (AMTB).}
\label{tab:amtb}
\centering
\setlength{\tabcolsep}{1.5mm}{
\begin{tabular}{c|ccc|cc|cc}
\hline
\multirow{2}{*}{Module}
&\multicolumn{3}{c|}{DepthTrack}
&\multicolumn{2}{c|}{RGBT234}
&\multicolumn{2}{c}{VisEvent}\\
\cline{2-8}
&Pr. &Re. &F &MPR &MSR &Pr. & AUC   \\
\hline
MCP* &60.6               &59.6     &60.1     &84.1     &61.5     &76.4    &59.3     \\
MCP &60.2               &59.0     &59.6     &83.7     &61.2     &75.9    &58.9     \\
LoRA &60.1               &59.0     &59.5     &84.1     &61.5     &76.1    &59.1      \\
FixedRank &60.5               &60.3     &60.4     &83.4     &61.2     &77.0    &59.9      \\
AMTB* &62.6               &61.9     &62.2     &86.4     &64.0     &77.6    &60.2      \\
\rowcolor{mygray}
AMTB &\textbf{62.9}               &\textbf{62.2}     &\textbf{62.5}     &\textbf{86.6}     &\textbf{64.1}     &\textbf{78.0}    &\textbf{60.7}      \\
\hline
\end{tabular}
}
\end{table}

\begin{table}[t]
\caption{Analysis of the rank budget of AMTB.}
\label{tab:rankbudget}
\centering
\setlength{\tabcolsep}{1.2mm}{
\begin{tabular}{c|c|c|ccc|cc|cc|c}
\hline
\multirow{2}{*}{index}
&\multirow{2}{*}{$\hat{n}$}
&\multirow{2}{*}{$\hat{m}$}
&\multicolumn{3}{c|}{DepthTrack}
&\multicolumn{2}{c|}{RGBT234}
&\multicolumn{2}{c|}{VisEvent} 
&\multirow{2}{*}{FPS} \\
\cline{4-10}
& & &Pr. &Re. &F &MPR &MSR &Pr. &AUC   \\
\hline
1&8  & 16  &61.8&61.3&61.6&85.8&63.5&76.7&59.7 &50    \\
2&16 & 16  &62.0&61.6&61.8&86.1&63.7&77.2&60.1 &48    \\
\rowcolor{mygray}
3&16 & 32  &\textbf{62.9}&\textbf{62.2}&\textbf{62.5}&\textbf{86.6}&\textbf{64.1}&\textbf{78.0}&\textbf{60.7}  &48   \\
4&16 & 48 &62.4&62.2&62.3&86.0&63.8&77.5&60.2  &48   \\
5&24 & 48 &62.9&61.7&62.1&86.2&64.0&77.8&60.4  &47   \\
\hline
\end{tabular}
}
\vspace{-3mm}
\end{table}

\noindent
\textbf{Effectiveness of the Auxiliary Modality Tuning Block.}
We conduct experiments on the multi-modal fusion module in Table~\ref{tab:amtb}.
$*$ means fine-tuning each video modality individually as ViPT does.
$MCP$ means that we use the MCP block of ViPT to fuse auxiliary video modalities.
$LoRA$ means that we add features of auxiliary video modalities to the image features after patch embedding and utilize the LoRA~\cite{lora} technique to fine-tune the foundation tracker.
$FixedRank$ means that we fix singular values in the auxiliary modality tuning block (AMTB) to 1.
Other training settings remain consistent with UniSOT.
Our designed AMTB outperforms $MCP$ and $LoRA$, which proves the effectiveness of the fusion designing.
Meanwhile, AMTB surpasses $MCP$ and $FixedRank$ with a large margin and outperforms AMTB with video modality-separate fine-tuning ($AMTB*$).
%
This is because AMTB constructs a compact modality-shared parameter space via modality-shared rank allocation that facilitates video modality-aligned feature learning. 
Different auxiliary video modalities benefit from joint fine-tuning with modality-shared parameters to learn more generalized features.
Meanwhile, AMTB adaptively assigns different ranks across video modalities via modality-specific rank allocation, enabling specialized feature learning while mitigating overfitting.

\noindent
\textbf{Analysis of the Rank Budget.} 
We study the effect of rank budget of AMTB in Table~\ref{tab:rankbudget}.
Some concepts need to be declared.
The inherent rank $r$ is set to 32.
$\hat{n}$ is the block-average modality-shared rank budget and $\hat{m}$ is the block-average modality-specific rank budget as stated in Section~\ref{sec:4.1.3}.
Comparing (1), (3) and (5), we can find that low $\hat{n}$ and $\hat{m}$ are not enough to support the tracker to learn effective features, while high $\hat{n}$ and $\hat{m}$ may lead to overfitting on the training set.
Comparing (2), (3) and (4), we can find that a proper ratio of $\hat{n}$ and $\hat{m}$ is beneficial to the tracker performance.
The reason is that low $\hat{m}/\hat{n}$ is not conducive to feature learning of modal alignment, while high $\hat{m}/\hat{n}$ limits the ability to learn video modality-specific features.
Moreover, larger $\hat{n}$ will reduce inference efficiency.
So we select $\hat{n}=16$ and $\hat{m}=8$ to keep good performance and efficiency.

\begin{table}[t]
\caption{Analysis of Incremental Video Modality Learning.}
\label{tab:incremental-r2}
\centering
\setlength{\tabcolsep}{1.5mm}{
\resizebox{\linewidth}{!}{\begin{tabular}{c|ccc|cc|cc}
\hline
\multirow{2}{*}{Module}
&\multicolumn{3}{c|}{DepthTrack}
&\multicolumn{2}{c|}{RGBT234}
&\multicolumn{2}{c}{VisEvent}\\
\cline{2-8}
&Pr. &Re. &F &MPR &MSR &Pr. & AUC \\
\hline
\rowcolor{mygray}
UniSOT-B (RGB+X) &{62.9}               &{62.2}     &{62.5}     &{86.6}     &{64.1}     &{78.0}    &{60.7}      \\
\hline
\textit{stage} 1: UniSOT-B (RGB+D) &62.6    &61.9     &62.2     &82.0     &61.3     &69.2    &51.9      \\
\textit{stage} 2: UniSOT-B (RGB+X) &62.9    &62.3     &62.6     &86.6     &64.1     &77.9    &60.7      \\
\hline
\textit{stage} 1: UniSOT-B (RGB+T) &51.8    &54.2     &53.1     &86.4     &64.0     &68.8    &51.3      \\
\textit{stage} 2: UniSOT-B (RGB+X) &62.8    &62.2     &62.5     &86.6     &64.1     &78.0    &60.6      \\
\hline
\textit{stage} 1: UniSOT-B (RGB+E) &50.6    &53.7     &52.1     &80.6     &60.3     &77.6    &60.2      \\
\textit{stage} 2: UniSOT-B (RGB+X) &62.8    &62.1     &62.4     &86.5     &64.1     &78.1    &60.8      \\
\hline
\end{tabular}}
}
\end{table}

\noindent
\textbf{Analysis of Incremental Video Modality Learning.}
We conduct further experiments to explicitly highlight RAMA’s effectiveness in enabling incremental video modality learning. 
Specifically, we implement a two-stage training process on UniSOT: in the first stage, we train UniSOT with RGB combined with a single non-RGB video modality, and in the second stage, we extend the training to RGB plus multiple non-RGB video modalities (denoted as X), including depth, thermal and event.
As detailed in Table~\ref{tab:incremental-r2}, the first-stage results demonstrate strong performance on evaluations specific to the trained video modality, but exhibit limited generalization to other video modalities. 
Crucially, after the second-stage training incorporating multiple video modalities, performance improved consistently across all video modality evaluations, regardless of the initial auxiliary video modality used in the first stage. 
This is because RAMA can adaptively allocate ranks among video modalities, thereby enabling incremental video modality learning.
These results intuitively highlight RAMA’s effectiveness in enabling incremental video modality learning.

\begin{table}[t]
\caption{Evaluation of UniSOT with Degraded RGB Inputs.}
\vspace{-1mm}
\label{tab:degrade}
\centering
\setlength{\tabcolsep}{1.5mm}{
\resizebox{\linewidth}{!}{
\begin{tabular}{c|ccc|cc|cc}
\hline
\multirow{2}{*}{Module}
&\multicolumn{3}{c|}{DepthTrack}
&\multicolumn{2}{c|}{RGBT234}
&\multicolumn{2}{c}{VisEvent}\\
\cline{2-8}
&Pr. &Re. &F &MPR &MSR &Pr. & AUC \\
\hline
UniSOT-B (Degraded RGB) &60.8               &58.4     &59.6     &85.3     &63.2     &75.3    &58.4      \\
\rowcolor{mygray}
UniSOT-B &\textbf{62.9}               &\textbf{62.2}     &\textbf{62.5}     &\textbf{86.6}     &\textbf{64.1}     &\textbf{78.0}    &\textbf{60.7}      \\
\hline
\end{tabular}
\vspace{-8mm}
}
}
\end{table}

\noindent
\textbf{Evaluation with Degraded RGB Inputs.}
We explored the model’s robustness to degraded RGB inputs by applying contrast reduction to RGB data in multi-modal video datasets.
Results in Table~\ref{tab:degrade} demonstrate that UniSOT maintains comparable performance with degraded RGB inputs, which validates its ability to utilize extra cues of auxiliary video modalities for robust tracking.
This is because RAMA allows the model to learn generalized representations from shared parameters and adaptively learn features crucial for different video modalities.

\subsection{Visualization}
\label{visualization}

\begin{figure}[t]
    \centering\includegraphics[width=\linewidth]{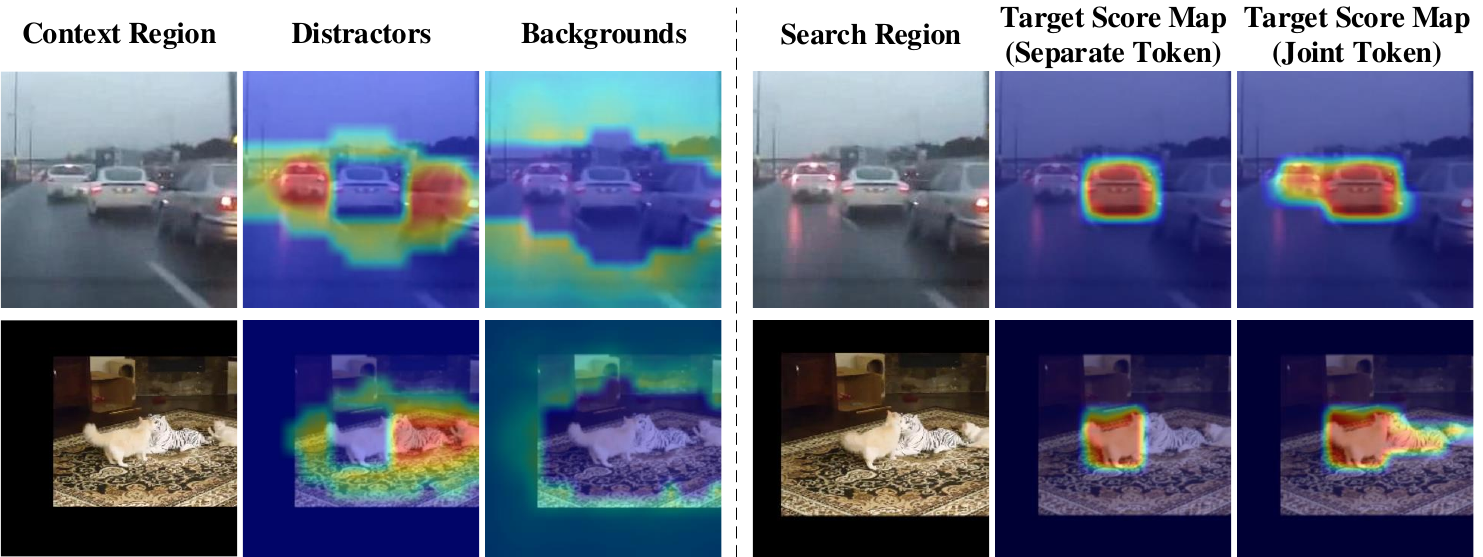}
    \vspace{-5mm}
    \caption{The left images show the distractor and background attention maps on the context region. The right images show the target score maps generated by separate distractor and background tokens or joint distractor and background token ($\beta=0.0$).}
    \label{fig:heatmap}
    \vspace{-3mm}
\end{figure}

\begin{figure*}[t]
    \centering
    \includegraphics[width=\linewidth]{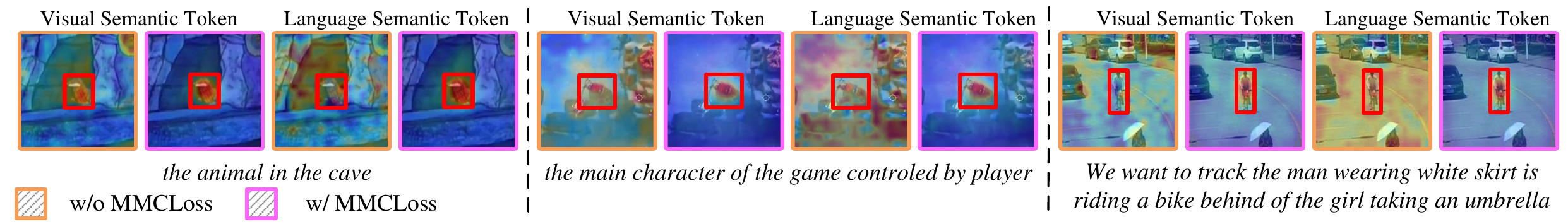}
    \vspace{-6mm}
    \caption{
    The response maps of visual and language semantic tokens in the search region under both settings, with and without MMCLoss.
    }
    \label{fig:mmcloss_res}
    \vspace{-3mm}
\end{figure*}

\begin{figure*}[t]
    \centering
    \includegraphics[width=1.0\linewidth]{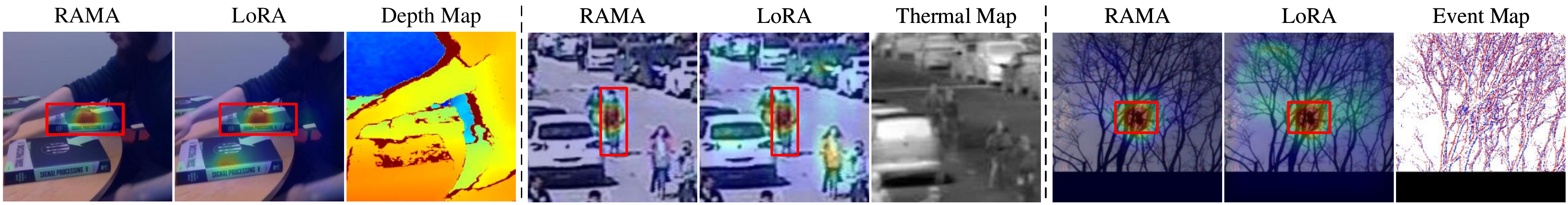}
    \vspace{-6mm}
    \caption{
    The classification maps of UniSOT with RAMA and LoRA in RGB+Depth, RGB+Thermal and RGB+Event datasets.
    }
    \vspace{-3mm}
    \label{fig:rama_vis}
\end{figure*}

\begin{figure}[t]
    \centering
    \includegraphics[width=0.6\linewidth]{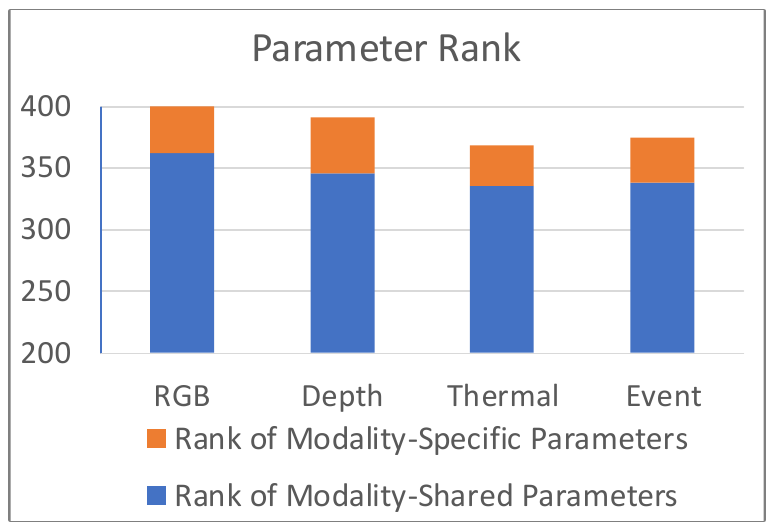}
    \vspace{-2mm}
    \caption{
    The parameter rank distribution among video modalities in AMTBs.
    }
    \label{fig:rama_stat}
\end{figure}

%

\textbf{Attention Maps and Target Score Maps}.
We visualize distractor attention maps and background attention maps in the distribution-based cross-attention.
As shown in Figure~\ref{fig:heatmap}, our designed distribution-based cross-attention can effectively divide out-box features into distractors and other backgrounds.
By such design, the extracted distractor token can retain more detailed information to distinguish subsequent distractors, improving the robustness of target localization.
Further, we visualize target score maps generated by separate distractor and background tokens or joint distractor and background token.
Here, ``separate" means we aggregate distractor and background features into different tokens, ``joint" means we aggregate all out-box features into one token ($\beta=0$ as discussed in Section~\ref{sec:4.3}).
The comparison of target score maps also shows that separate distractor and background tokens can obtain more accurate localization results.
These visualizations demonstrate the effectiveness of our proposed distribution-based cross-attention mechanism.

\noindent
\textbf{Response Map of Semantic Tokens}.
We further provide response maps of visual and language semantic tokens in the search region under both settings, with and without MMCLoss, as shown in Figure~\ref{fig:mmcloss_res}. 
In the absence of MMCLoss, two phenomena can be observed: first, the response distributions of visual and language semantic tokens exhibit inconsistency, reflecting the semantic gap between vision and language features. 
During multi-modal joint training, this inconsistency may cause the model to optimize in divergent directions for different reference modalities, resulting in compromised performance. 
Second, both visual and language semantic tokens demonstrate limited discriminability in the search region, reflecting the poor robustness of the reference features.
Conversely, with MMCLoss, visual and language semantic tokens achieve consistent and discriminative responses, indicating that MMCLoss aligns vision and language features into a unified semantic space and facilitates consistent feature learning across reference modalities while enhancing the discriminability of reference features.

\noindent
\textbf{Center Score Maps with AMTB and LoRA}.
Comparative visualizations of center score map with AMTB or LoRA in Figure~14 demonstrate that our approach achieves more stable localization across auxiliary video modalities, as shown in Figure~\ref{fig:rama_vis}. 
This is because AMTB adaptively distributes parameters to simultaneously capture video modality-shared and video modality-specific features as shown in Figure~\ref{fig:rama_stat}, thereby improving multi-modal localization robustness through joint fine-tuning.

\subsection{Applications \& Limitations}
\textbf{Applications.}
{Drones or professional recording equipment ($e.g.$, camera gimbal) typically use single-object tracking algorithms to continuously estimate the location of the target object during automatic tracking. 
In practice, users may prefer various reference modalities to specify the target object for convenience, while the demand for different video modalities varies across scenarios to enhance tracking robustness.
The generalized capabilities of UniSOT provide users with significant convenience and functional flexibility within a single model, eliminating the need for multiple specialized systems.
Additionally, we believe that the exploration of general capability models also potentially drives the development of artificial general intelligence (AGI).}

\noindent\textbf{Limitations}
{As highlighted, UniSOT focuses on establishing a unified multi-modal tracking framework with strong generalization across benchmarks.
Considering the simplicity of framework, UniSOT does not explicitly model the issues of target disappearance and re-detection, which remain an open challenge for future research.
In practical applications, UniSOT can integrate off-the-shelf global detection modules~\cite{globaltrack} to address target disappearance/reappearance scenarios ($e.g.$, out of view), as STARK-LT~\cite{vot2021} does.}

%% file: conclusion.tex
In this work, we propose a novel unified tracking framework for different combinations of reference and video modalities. 
To the best of our knowledge, UniSOT is the first tracker to support tracking
with three reference modalities (NL, BBOX, NL+BBOX) in sequences of four video modalities (RGB, RGB+Depth, RGB+Thermal, RGB+Event) under a unified architecture with uniform parameters, enabling more generalized application scenarios.
Extensive experimental results on visual tracking, vision-language tracking, grounding and RGB-X (depth, thermal, event) tracking datasets demonstrate that UniSOT shows superior results against modality-specific counterparts.

%% file: sup.tex
\begin{table}[!t]
\caption{Analysis of the model training strategy of the first stage.}
\label{tab:training_strategy}
\centering
\setlength{\tabcolsep}{1.5mm}{
\begin{tabular}{c|c|cc|cc|cc}
\hline
\multirow{3}{*}{Pretrain}
&\multirow{3}{*}{Ratio}
&\multicolumn{6}{c}{TNL2K}\\
\cline{3-8}
&&\multicolumn{2}{c|}{BBOX}
&\multicolumn{2}{c|}{NL}
&\multicolumn{2}{c}{NL+BBOX} \\
\cline{3-8}
&&\makebox[1pt][c]{AUC} &P &\makebox[1pt][c]{AUC} &P &\makebox[1pt][c]{AUC} &P   \\
\hline
\makebox[45pt][c]{BERT+MAE} &3:1:3               &60.6&61.5&53.8&54.3&60.8&62.5     \\
\rowcolor{mygray}
\makebox[45pt][c]{BERT+MAE} &4:1:4 &              60.9&61.9&\textbf{53.9}&\textbf{54.4}&61.2&62.8     \\
\makebox[45pt][c]{BERT+MAE} &5:1:5   &            \textbf{61.0}&\textbf{62.1}&53.4&53.7&\textbf{61.3}&\textbf{63.0}     \\
MAE      &4:1:4               &60.4&60.7&52.3&51.8&60.1&61.5     \\
\hline
\end{tabular}
}
\end{table}

\begin{table}[!t]
\caption{Analysis of the update interval of scenario tokens.}
\label{tab:update_interval}
\setlength{\tabcolsep}{1.5mm}{
\begin{center}
\begin{tabular}{c|cc|cc|cc}
\hline
\multirow{3}{*}{update interval}
&\multicolumn{6}{c}{TNL2K}\\
\cline{2-7}
&\multicolumn{2}{c|}{BBOX}
&\multicolumn{2}{c|}{NL}
&\multicolumn{2}{c}{NL+BBOX} \\
\cline{2-7}
&AUC &P &AUC &P &AUC &P   \\
\hline
$\infty$         &60.3&61.1&52.4&52.2&60.2&61.5  \\
50               &60.8&61.8&53.4&53.9&61.0&62.5     \\
\rowcolor{mygray}
20               &\textbf{60.9}&\textbf{61.9}&\textbf{53.9}&\textbf{54.4}&\textbf{61.2}&\textbf{62.8}     \\
10               &\textbf{60.9}&61.8&53.7&54.1&61.1&62.5     \\
\hline
\end{tabular}
\end{center}
}
\vspace{-4mm}
\end{table}

\begin{table}[t]
\caption{Analysis of Single Video Modality Input.}
\label{tab:single}
\centering
\setlength{\tabcolsep}{1.5mm}{
\begin{tabular}{c|ccc|cc|cc}
\hline
\multirow{2}{*}{Module}
&\multicolumn{3}{c|}{DepthTrack}
&\multicolumn{2}{c|}{RGBT234}
&\multicolumn{2}{c}{VisEvent}\\
\cline{2-8}
&Pr. &Re. &F &MPR &MSR &Pr. & AUC \\
\hline
UniSOT-B* (RGB) &52.7               &54.9     &53.8     &82.2     &61.6     &69.6    &52.6      \\
UniSOT-B (RGB) &53.2               &55.2     &54.2     &82.4     &61.7     &69.8    &52.7      \\
UniSOT-B (X) &26.4    &17.0     &20.7     &28.5     &20.8     &23.8    &14.6      \\
\rowcolor{mygray}
UniSOT-B (RGB+X) &\textbf{62.9}               &\textbf{62.2}     &\textbf{62.5}     &\textbf{86.6}     &\textbf{64.1}     &\textbf{78.0}    &\textbf{60.7}      \\
\hline
\end{tabular}
}
\end{table}

\begin{figure*}[t]
    \centering\includegraphics[width=1.0\linewidth]{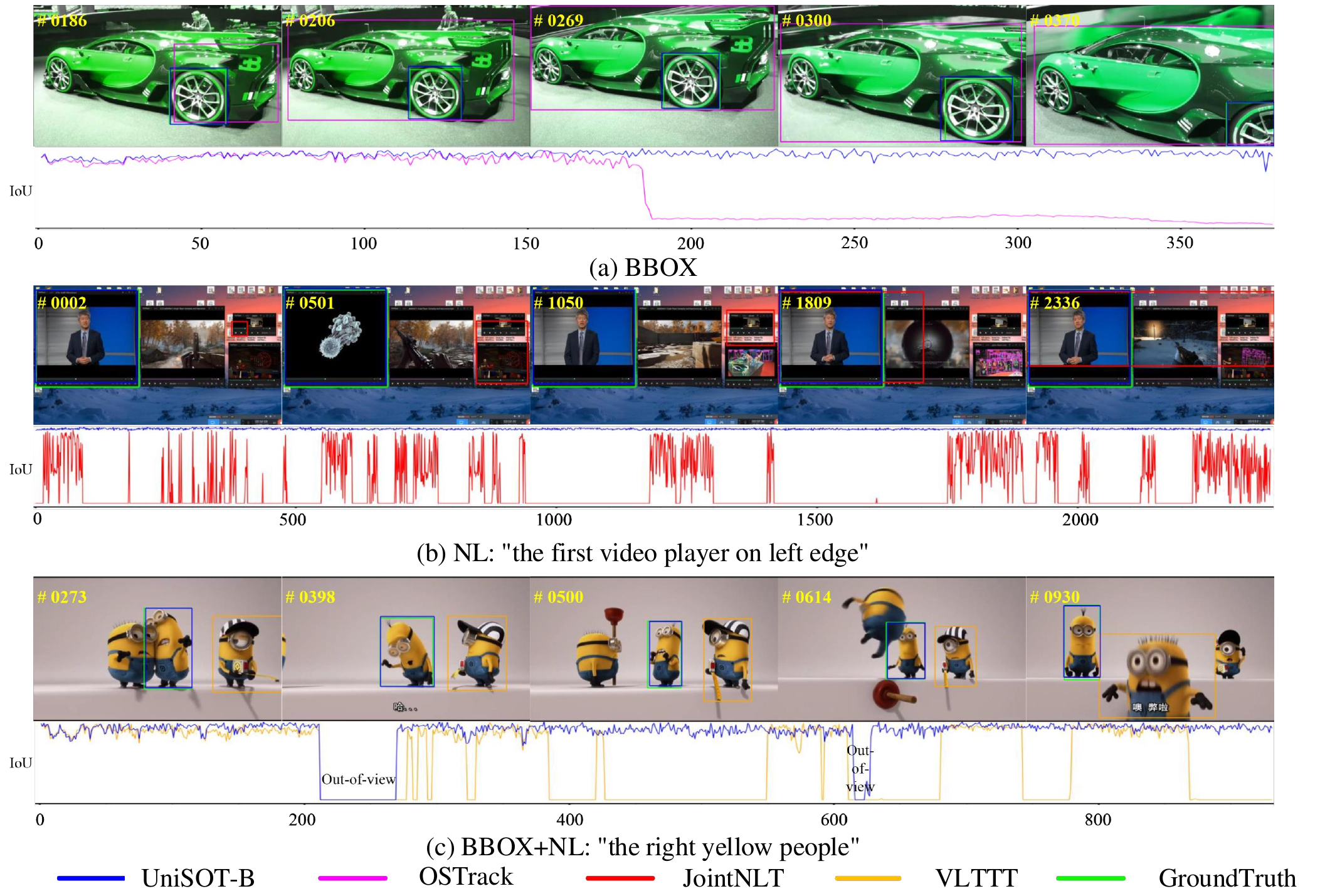}
    \caption{Visualization of tracking results with different reference modalities. UniSOT achieves more robust tracking performance compared with modality-specific counterparts, which proves the generalization of UniSOT.}
    \label{fig:tracking_results}
\end{figure*}

\begin{figure*}[tp]
    \centering\includegraphics[width=1.0\linewidth]{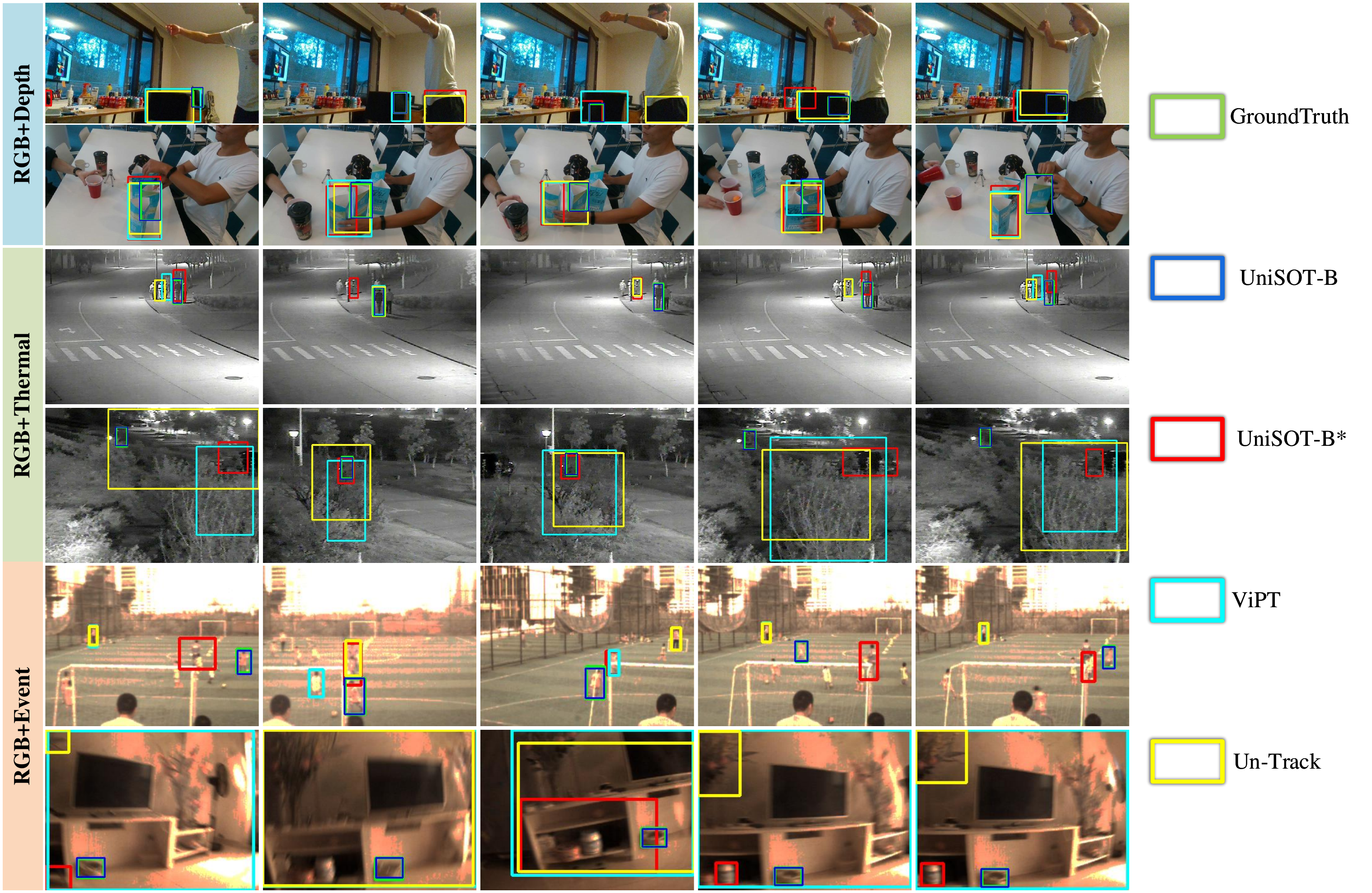}
    \vspace{-3mm}
    \caption{Visualization of tracking results with different auxiliary modalities. UniSOT achieves more robust tracking performance compared with previous state-of-the-art trackers.}
    \label{fig:uvltrackX-tracking_results}
\end{figure*}

\begin{figure*}[tp]
    \centering\includegraphics[width=1.0\linewidth]{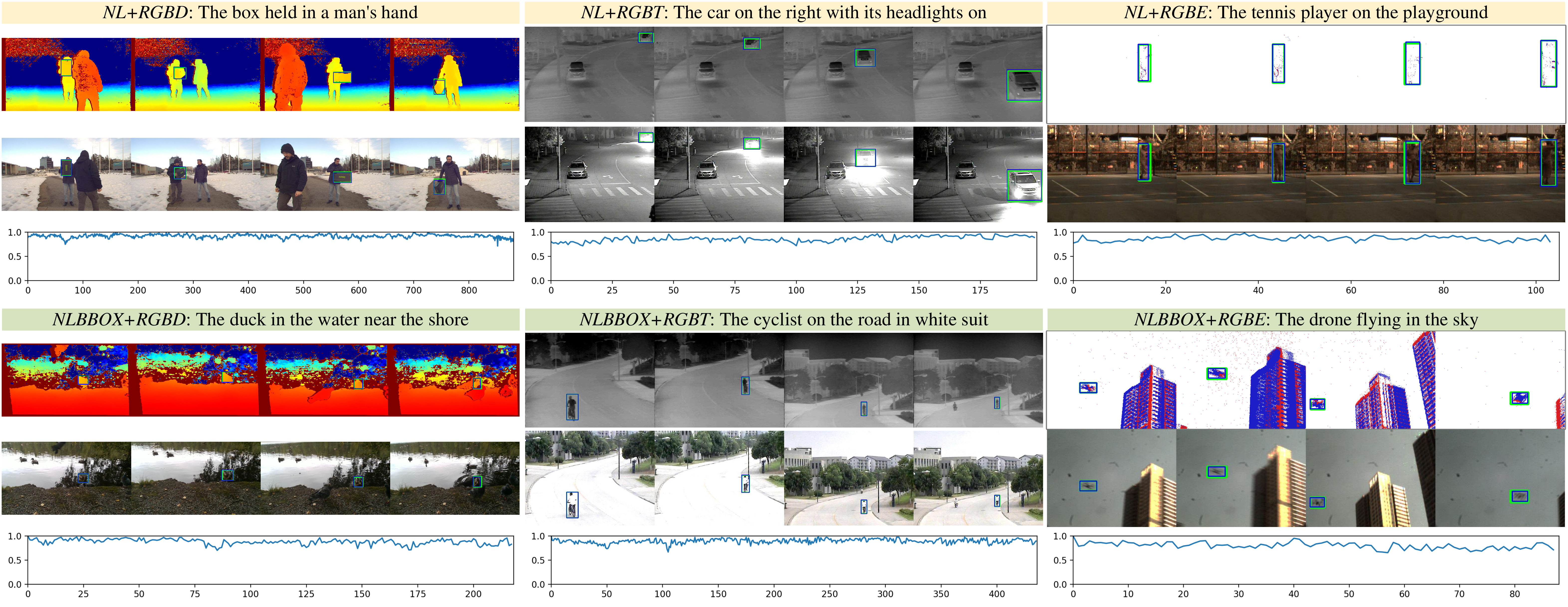}
    \caption{Tracking with different combinations of reference and video modalities. The green box is groundtruth, blue box is the predicted box.}
    \label{fig:nl_rgbx}
\end{figure*}

\subsection{Extra Ablation Study}

\noindent
\textbf{Analysis of the Training Strategy of the First Stage.}
As shown in Table~\ref{tab:training_strategy}, the ratio of different references (BBOX:NL:Both) has little effect on the performance of UniSOT.
Moreover, we try to initialize all parameters in the backbone with ViT parameters pretrained by MAE, which reduces overall performance.
This is because pretrained BERT parameters bring initial language modeling capabilities to the tracker to understand natural language.

\noindent
\textbf{Analysis of the Update Interval.}
The scenario tokens in the reference-adaptive box head are updated at regular intervals, which reduces computational costs and avoids error accumulation.
As shown in Table~\ref{tab:update_interval}, the performance is insensitive to the update interval in a certain range.
If scenario tokens are not updated, the performance decreases overtly.
It demonstrates that dynamic scenario cues are vital for robust tracking.

\noindent
\textbf{Analysis of Single Video Modality Input.}
We conduct experiments to evaluate the performance of RGB-only inputs, UniSOT-B (RGB), and the auxiliary video modality-only inputs, UniSOT-B (X), on the RGB+X tracking setup as shown in Table~\ref{tab:single}.
UniSOT-B* means UniSOT only trained with the first training stage.
As we can see, UniSOT-B achieves a slight performance improvement on RGB inputs after the second training stage (fine-tuning with AMTB on RGB+X dataset).
These results demonstrate the robustness of the unified architecture when not using the auxiliary video modality.
The performance drops significantly when only auxiliary video modality are used, as these modalities lack discriminative texture information, making target localization challenging. 
When both RGB and auxiliary video modalities serve as inputs, there is a notable performance boost. 
This proves its capability to effectively leverage multi-modal information for enhanced tracking performance.

\subsection{Visualization}

\noindent
\textbf{Tracking Results}.
We visualize some tracking results with different reference modalities.
As shown in Figure~\ref{fig:tracking_results}(a), compared with OSTrack~\cite{OSTrack}, UniSOT can achieve more robust tracking in the complex scenario with the bounding box reference.
This is because our designed reference-adaptive box head can utilize dynamically mined scenario cues to discriminate target object, reducing the impact of complex backgrounds.
Meanwhile, as shown in Figure~\ref{fig:tracking_results}(b), UniSOT has excellent visual grounding ability, thus achieving more accurate localization results compared with JointNLT~\cite{JointNLT} using the natural language reference.
Moreover, for tracking by language and box specification, VLTTT~\cite{VLT} suffers from frequent target drift, while UniSOT can maintain high-quality tracking results in long-term video, even after target object disappears and reappears.
It benefits from modality-aligned feature learning, which can make full use of different modal references to enhance target features, so as to realize more robust tracking.
Further, as shown in Figure~\ref{fig:uvltrackX-tracking_results}, UniSOT achieves robust tracking results in challenging scenarios compared with UniSOT and previous state-of-the-art RGB-X trackers.
This proves the advantage of the rank-adaptive designing that enables both modality-aligned and modality-specific feature learning.
The above results intuitively confirm that our proposed framework can achieve superior performance across combinations of reference and video modalities.

\noindent
\textbf{More combinations of reference and video modalities.}
We further explore the combination of different reference and video modalities.
As shown in Figure~\ref{fig:nl_rgbx}, UniSOT can achieve superior performance under different combinations of reference and video modalities.
This proves that UniSOT can track with different reference modalities in sequences of different video modalities.
UniSOT not only can simultaneously cope with three types of target references, enabling various human-machine interactions, but also can freely integrate different auxiliary modalities to enhance robustness in complex scenarios.